%% file: main.tex
\documentclass[runningheads]{llncs}

\usepackage{eccv}

\usepackage{eccvabbrv}

\usepackage{graphicx}
\usepackage{booktabs}

\usepackage[accsupp]{axessibility}  

\usepackage[colorlinks,citecolor=eccvblue,linkcolor=red,urlcolor=magenta]{hyperref}

\usepackage{orcidlink}
\usepackage{fontawesome5}

\usepackage{algorithmic}
\usepackage{algorithm}
\usepackage{float}
\usepackage{graphicx}

\usepackage{booktabs}
\usepackage{multirow}
\usepackage{array}
\usepackage{colortbl}
\usepackage{xcolor}

\usepackage{makecell}
\usepackage{amssymb}
\usepackage{subcaption}
\usepackage{wrapfig}

\usepackage{adjustbox}      
\usepackage{threeparttable} 

\AtEndPreamble{
  \crefname{figure}{Fig.}{Figs.}
  \Crefname{figure}{Figure}{Figures}
  \crefname{equation}{Eq.}{Eqs.}
  \Crefname{equation}{Equation}{Equations}
  \crefname{algorithm}{Alg.}{Algs.}
  \Crefname{algorithm}{Algorithm}{Algorithms}
}

\definecolor{bestblue}{HTML}{E3F2FD}
\definecolor{hpdblue}{HTML}{1565C0}
\definecolor{rowgray}{HTML}{F5F5F5}
\definecolor{lightred}{HTML}{FFEBEE}
\definecolor{highlight}{gray}{0.93}

\makeatletter
\renewcommand\@fnsymbol[1]{\ensuremath{\ifcase#1\or\dagger\or\ddagger\or\mathsection\or\mathparagraph\or\|\else\@ctrerr\fi}}
\makeatother
\begin{document}

\title{Reward Lightning: Fast Video Generation via Homologous Preference Distillation} 

\titlerunning{Reward Lightning}

\author{Jiaxiang Cheng$^{*\dagger}$ \and Bing Ma \and Xuhua Ren \and Kai Yu \and Peng Zhang \\
Tianxiang Zheng \and Qinglin Lu}

\authorrunning{J.~Cheng et al.}

\institute{Tencent Hunyuan, China\\
\email{jiaxiangcc@gmail.com}}

\maketitle
\let\thefootnote\relax\footnotetext{$^{*}$Corresponding Author.\quad $^{\dagger}$Project Lead.}

\input{00_abstract}
\input{01_introduction}

\input{02_related_work}
\input{03_methodology}
\input{04_experiment}
\input{06_conclusion}

%
%
\bibliographystyle{splncs04}
\bibliography{main}

\clearpage
\appendix
\input{07_appendix}

\end{document}

%% file: 00_abstract.tex
\begin{abstract}

Achieving simultaneous preference alignment and distillation acceleration in video diffusion models remains an open challenge.
Existing methods optimize the two objectives over mismatched representation spaces, where improving one objective often compromises the other.
To overcome this, we propose \textbf{Reward Lightning}, a unified framework that aligns and accelerates a video diffusion model within a single shared representation.
Its central principle is \emph{homology}: both objectives are evaluated on identical latent features, which mitigates the gradient conflicts that arise when they are optimized over disjoint representations.
As a foundational component, we first introduce a latent reward model (LRM) that scores videos directly in the latent space, without decoding back to the pixel space.
Building on the LRM, homologous preference distillation (HPD) reuses this shared backbone to perform adversarial distillation and preference alignment jointly, yielding few-step generators that remain faithful and well aligned.
Extensive experiments demonstrate that the LRM surpasses pixel-level and latent-level reward baselines by $11.0\%$ and $14.7\%$ in preference accuracy, and that Reward Lightning generates high-fidelity videos in merely $1$ to $4$ steps, improving the average VBench score by $2.1\%$ while leading in text alignment, motion quality, and visual quality.
Project page: \url{https://reward-lightning.github.io}.

  \keywords{Video Diffusion Models \and Preference Alignment \and  Distillation Acceleration}
\end{abstract}

%% file: 01_introduction.tex
\section{Introduction}
\label{introduction}

Video diffusion models~\cite{singer2023makeavideo,guo2024animatediff,ho2022video,blattmann2023align,wan2025wan,kong2024hunyuanvideo} have substantially advanced visual generation.
Their large parameter counts, long temporal sequences, and iterative sampling, however, incur prohibitive inference costs.
Distillation techniques~\cite{wang2024phased,yin2023one,lin2025diffusion,salimans2022progressive,song2023consistency,geng2025mean,zhang2025turbodiffusion,zheng2026large} effectively reduce the number of sampling steps, yet they inevitably distort the generative priors essential for reflecting human preferences~\cite{xu2023imagereward,lee2023aligning,black2024training,wallace2024diffusion,liu2025improving,xu2024visionreward}.
This raises a key question: how can we integrate preference alignment into the distillation trajectory without compromising fidelity?

\input{figures/main_1}

Recent methods such as DMDR~\cite{jiang2025distribution}, FlashDMD~\cite{chen2025flash}, and DOLLAR~\cite{ding2025dollar} integrate reinforcement learning with distillation to achieve aligned few-step generation.
These paradigms, however, rely on heterogeneous optimization structures: they either linearly combine a pixel-space reward model with latent-space distillation, or update disjoint parameters for the two objectives (as shown in Fig.~\hyperref[fig:main_1]{\ref*{fig:main_1}-(a,b)}).
In the context of multi-objective optimization, such structural and representational heterogeneity aggravates gradient conflicts, particularly in the high-noise regimes of few-step sampling, where reaching an optimal equilibrium between generation fidelity and human preference becomes difficult.

To overcome this bottleneck, we propose \textbf{Reward Lightning}, a unified framework built on the principle of \textit{homology}, strict structural and representational consistency between preference alignment and distillation.
Because both objectives are evaluated on identical latent features, homology removes the representational mismatch underlying the gradient conflicts above.
Reward Lightning realizes this principle through two components: a Latent Reward Model (LRM) that supplies the shared backbone, and Homologous Preference Distillation (HPD) that learns over it.
Recent work like PRFL~\cite{mi2025video} has elegantly shown that pre-trained video diffusion models are naturally suited to process-aware reward modeling in the latent space.
We share this latent-evaluation philosophy, but design our LRM to play a dual role: not only as a standalone optimization target, but also as a structural prerequisite for preference distillation.
To evaluate the highly blurred trajectory of few-step generation ($\le 4$ NFEs), we augment the preference dataset with few-step samples alongside their multi-step counterparts.
Unlike existing methods~\cite{lin2025diffusion,cheng2026phased,mi2025video} that apply time-step conditioning to the video representations alone while relying on a purely trainable query, we project the time-step embeddings directly as the dynamic query, allowing it to adaptively capture the varying representations along the flow trajectory.

The gradient conflict diagnosed above could in principle be resolved by explicit multi-objective optimization, but such treatment is computationally prohibitive for large-scale flow matching models.
Building on the shared LRM backbone, HPD mitigates the conflict through a unified design rather than resolving it explicitly.
By enforcing \textit{structural homology} (a shared reward backbone) and \textit{space homology} (the same latent representations), both the distillation and preference gradients are computed within an identical latent manifold.
This homologous constraint acts as an implicit gradient regularizer: it promotes alignment between the gradient directions and stabilizes the joint optimization.
In this way, the homology integrates preference alignment into the distillation trajectory, yielding few-step samples of both high fidelity and strong human alignment.

Extensive experiments validate the effectiveness of our unified framework.
As a structural foundation, our LRM attains a preference accuracy on VideoGen-RewardBench~\cite{liu2025improving} that surpasses existing pixel-level and latent-level baselines by 11.0\% and 14.7\%, respectively.
Powered by this homology, our HPD reduces the sampling process to $1-4$ NFEs, as evaluated on VBench~\cite{huang2024vbench}.
Under this few-step setting, our approach achieves a superior balance between inference efficiency and generation quality compared with both pure distillation and preference distillation methods.
Specifically, it improves the average overall score by 2.1\% and sets a new record in text alignment, motion quality, and visual quality.

%% file: figures/main_1.tex
\begin{figure}[tb]
  \centering
  \includegraphics[width=0.95\textwidth]{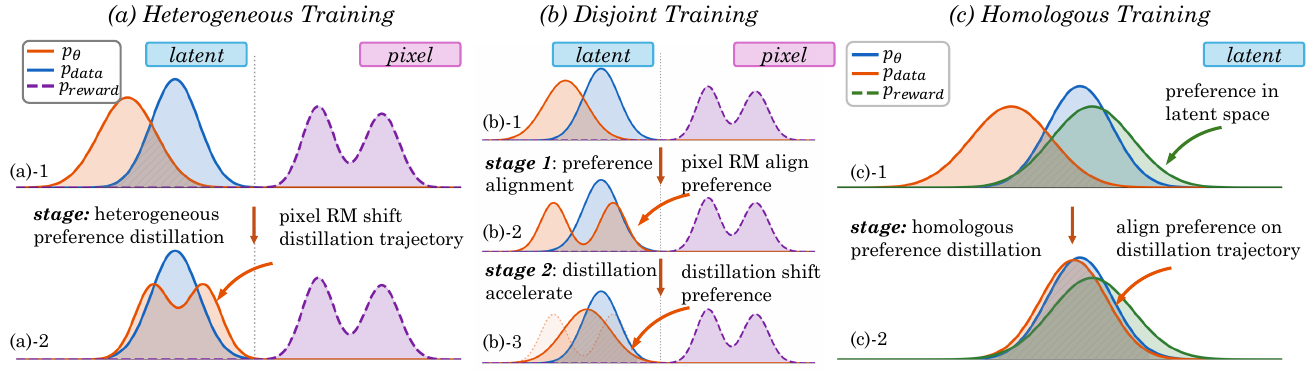}  
  \caption{
  \textbf{Motivation}.
  \textit{(a) Heterogeneous Training}: Pixel rewards induce shifts in distillation trajectories through multi-objective gradient conflicts.
  \textit{(b) Disjoint Training}: Sequential training discards original preference distributions through catastrophic forgetting.
  \textit{(c) Homologous Training}: Preference distillation based on homologous structures and data guides the generator toward the jointly optimal distribution.
}
  \label{fig:main_1}
\end{figure}

%% file: 02_related_work.tex
\section{Related Work}
\label{related-work}

\subsubsection{Post-Training.}
Post-training of diffusion models pursues two largely separate goals: preference alignment and distillation acceleration.
To align outputs with human preferences, Reinforcement Learning from Human Feedback (RLHF)~\cite{christiano2017deep,ouyang2022training,bai2022training} has been adapted to diffusion models, either through learned reward models~\cite{xu2023imagereward,kirstain2023pick} or Direct Preference Optimization (DPO)~\cite{rafailov2023direct,wallace2024diffusion,yang2024using}.
These alignment methods, however, typically operate on multi-step sampling~\cite{ho2020denoising,song2021scorebased}.
Distillation, by contrast, targets the high inference latency of iterative sampling.
Progressive distillation~\cite{salimans2022progressive,meng2023distillation,lin2024sdxl}, consistency models~\cite{song2023consistency,luo2023latent,kim2024consistency,wang2024phased}, distribution matching~\cite{yin2023one,yin2024improved}, and adversarial distillation~\cite{sauer2024adversarial,sauer2024fast,lin2025diffusion,cheng2026phased} all compress the generation trajectory into a few steps.
Yet these methods prioritize fidelity preservation and distribution matching, and largely overlook human preference alignment.
As a result, alignment and acceleration have advanced largely in isolation.

\subsubsection{Joint Preference Optimization and Distillation.}
Recent methods~\cite{jiang2025distribution,chen2025flash,eyring2025noise,zhang2024unifl,ding2025dollar} attempt to integrate preference optimization with distillation, so as to attain efficiency and alignment simultaneously~\cite{sauer2024fast,wallace2024diffusion}.
Sequential approaches address the two tasks in disjoint stages, but the second stage inevitably shifts the distribution learned in the first, a manifestation of catastrophic forgetting~\cite{french1999catastrophic,kirkpatrick2017overcoming}.
Naive joint training, in turn, linearly combines a pixel-level preference reward with a latent-level distillation objective; this cross-domain mismatch induces severe gradient conflicts, as pixel-space optimization directly interferes with latent-space distillation.
Neither paradigm maximizes both acceleration and alignment.
The root cause is representational: rewards and distillation are optimized in mismatched spaces, which motivates aligning them within a single representation.

%% file: 03_methodology.tex
\section{Methodology: Reward Lightning}
\label{methodology}

In this section, we propose Reward Lightning, a unified framework for simultaneous preference alignment and distillation acceleration in flow matching models.
Its central concept is homology:
by enforcing structural and space homology, both the distillation and preference objectives are evaluated within a shared latent representation space.
Below, we first review preference alignment and adversarial distillation (\cref{preliminaries}); then introduce our latent reward model, which serves as the structural foundation (\cref{latent-reward-model}); and finally detail the mechanism and optimization strategy of our framework (\cref{homologous-preference-distillation}).

\input{figures/main_2}

\subsection{Preliminaries: Brief Review for Preference and Distillation}
\label{preliminaries}

\subsubsection{Reinforcement Learning from Human Feedback.}
\label{reinforcement-learning-from-human-feedback}

Reinforcement Learning from Human Feedback (RLHF)~\cite{lee2023aligning,wu2023human,black2024training,xu2023imagereward,kirstain2023pick} aligns a pre-trained generator $v_\theta(\cdot,t)$ with human preferences, typically through a pixel-space reward model $R_\phi(\cdot)$.
Given a preference dataset $\mathcal{D}_\text{pair}=\{ (\mathbf{x}^\text{w},\mathbf{x}^\text{l}) \}$ comprising preferred videos $\mathbf{x}^\text{w}$ and rejected videos $\mathbf{x}^\text{l}$, $R_\phi(\cdot)$ is optimized via the Bradley-Terry (BT) objective~\cite{terry1925bt}:
\begin{equation}
    \mathcal{L}_\text{RM}(\phi) = - \mathbb{E}_{(\mathbf{x}^\text{w}, \mathbf{x}^\text{l}) \sim \mathcal{D}_\text{pair}} \left[ \log \sigma \left( R_\phi(\mathbf{x}^\text{w}) - R_\phi(\mathbf{x}^\text{l}) \right) \right],
\end{equation}
where $\sigma(\cdot)$ is the sigmoid function.
In standard latent-based generative frameworks~\cite{rombach2022high}, the optimization occurs in the latent space $\mathbf{z} \in \mathcal{Z}$, where $\mathbf{z} = g_\text{VAE}^E(\mathbf{x})$ is extracted via a pre-trained Variational Auto-Encoder (VAE) encoder~\cite{2014diederikAuto}.
To transfer human preferences to the generator $v_\theta(\cdot,t)$, Reward Feedback Learning (ReFL)~\cite{xu2023imagereward} directly maximizes the reward of the predicted clean flow state $\hat{\mathbf{z}}_0^\text{w,l} = f_\theta(\mathbf{z}_t^\text{w,l}, t)$, where $f_\theta(\mathbf{z}_t,t) := \mathbf{z}_t - t \cdot v_\theta(\mathbf{z}_t,t)$ represents the flow trajectory mapping toward $t=0$.
Because the reward model evaluates pixel-space representations, the ReFL objective is formulated with an explicit decoding step:
\begin{equation}
\mathcal{L}_\text{ReFL}(\theta) = - \mathbb{E}_{t\in(0,1]} \left[ R_\phi(g_\text{VAE}^D(f_\theta(\mathbf{z}_t,t))) \right].
\end{equation}

\subsubsection{Adversarial Distillation.}

Adversarial distillation accelerates iterative sampling by compelling the generator $v_\theta(\cdot,t)$ to map the noise prior $\mathbf{z}_1 \sim \mathcal{N}(0;\mathbf{I})$ directly to the real latent data distribution $p_\text{data}(\mathbf{z})$ in a few steps.
It introduces a trainable discriminator $D_\psi(\cdot,\cdot)$ to distinguish between real latent samples $\mathbf{z}_0$ and generated predictions $\hat{\mathbf{z}}_0$.
To evaluate representations across varying noise levels, intermediate states $\mathbf{z}_t$ and $\hat{\mathbf{z}}_t$ are constructed by perturbing $\mathbf{z}_0$ and $\hat{\mathbf{z}}_0$ with noise at timestep $t$.
Typically, the discriminator is trained via the standard Hinge loss~\cite{miyato2018spectral}:
\begin{equation}
    \mathcal{L}^D_\text{ADV}(\psi) = \frac{1}{2} \mathbb{E}_{\mathbf{z}_0 \sim p_\text{data}} \left[\max(0, 1 - D_\psi(\mathbf{z}_t,t))\right] + \frac{1}{2} \mathbb{E}_{\hat{\mathbf{z}}_0\sim p_\theta} \left[\max(0, 1 + D_\psi(\hat{\mathbf{z}}_t,t))\right].
\end{equation}
The generator is optimized to deceive the discriminator via the adversarial loss:
\begin{equation}
\mathcal{L}^G_\text{ADV}(\theta) = - \mathbb{E}_{\hat{\mathbf{z}}_0 \sim p_\theta} \left[D_\psi(\hat{\mathbf{z}}_t,t)\right].    
\end{equation}
This minimax formulation ensures that the generator produces high-fidelity structural distributions without relying on multi-step solver evaluations.

\subsection{Latent Reward Model: As a Structural Prerequisite}
\label{latent-reward-model}

\subsubsection{Multi-Margin Dataset Construction.}
\label{multi-margin-dataset-collection}

Constructing a generalizable reward model requires a dataset that spans diverse distributions.
As shown in Fig.~\hyperref[fig:main_2]{\ref*{fig:main_2}-a}, we curate a multi-margin dataset comprising preferred and rejected videos across distinct comparative scenarios (more annotation details in the appendix):

\begin{itemize}
\item \textit{Intra-Model Pairs.} 
We collect $30,000$ pairs generated by the target model across random seeds.
Five professional annotators established preferences based on visual quality, motion dynamics, and semantic alignment.
This subset drives self-refinement, steering the generator toward its optimal modes.
\item \textit{Inter-Model Pairs.} 
Sourced from VisionRewardDB~\cite{xu2024visionreward}, this subset contains $17,390$ expertly annotated pairs generated by five state-of-the-art models.
This subset prevents the reward model from overfitting to internal distributions, and enforces robust boundary generalization.
\item \textit{Real-Synthetic Pairs.} 
We construct $12,000$ pairs comparing real-world videos with generated videos.
We designate the real videos as preferred and the generated videos as rejected.
These pairs raise the generation quality ceiling.
\end{itemize}

\paragraph{Few-step Augmentation for Distillation.} 
Standard preference optimization evaluates high-fidelity multi-step samples.
However, preference distillation entails evaluating accelerated few-step predictions spanning $N\in\{1,2,3,4\}$ steps that often exhibit severe structural blurring and thus act as out-of-distribution data.
To mitigate this domain gap, we augment the dataset with $5,000$ specialized pairs.
We pair high-quality multi-step outputs as preferred targets with their structurally degraded few-step counterparts as rejected samples.
This targeted augmentation equips the reward model to provide reliable gradient guidance across highly truncated distillation trajectories.

\subsubsection{Reward Model Architecture.}
\label{reward-model-design}

As shown in Fig.~\hyperref[fig:main_2]{\ref*{fig:main_2}-b}, we initialize our LRM backbone $F_\phi(\cdot,t)$ with a pre-trained diffusion model~\cite{wan2025wan}.
Given a preference latent pair $(\mathbf{z}_0^\text{w}, \mathbf{z}_0^\text{l})$, we sample intermediate flow states $\mathbf{z}_t=(1-t)\mathbf{z}_0 + t\mathbf{z}_1$, where $t\sim \mathcal{U}(0,1)$ and $\mathbf{z}_1\sim \mathcal{N}(0;\mathbf{I})$.
The backbone extracts spatial features $F_t = F_\phi (\mathbf{z}_t, t)$.
This latent-space evaluation bypasses VAE decoding, ensuring preference and distillation gradients share an identical manifold.

\paragraph{Time-Modulated Attention Head.}

To evaluate representations across the severe distributional shifts of the flow trajectory, the reward head $R_\phi^\text{head}(\cdot,\cdot)$ must be noise-aware.
Unlike traditional architectures that rely solely on learnable queries and struggle to generalize across varying timesteps~\cite{mi2025video,cheng2026phased,lin2025diffusion}, we dynamically modulate the continuous flow state.
Specifically, for each of the $H{=}3$ attention heads $h$, we project the timestep embedding $\mathbf{e}_t$ via a weight matrix $W_h$ into a dynamic state query $\mathbf{s}_{t,h}$.
This time-dependent query adaptively aggregates the spatial features $F_t$ via cross-attention.
A linear projection $W_o$ then maps the concatenated outputs to the final scalar reward:
\begin{equation}
R_\phi^\text{head}(F_t, t) = W_o \operatorname{Concat}_{h=1}^{H} \big( \operatorname{CrossAttn}(\mathbf{s}_{t,h}, F_t) \big).
\end{equation}
This temporal modulation allows the head to shift its receptive focus—capturing macroscopic structural layouts in high-noise regimes and fine-grained textures in low-noise regimes—providing precise, state-aware gradient guidance.

\subsubsection{Reward Model Training}
\label{reward-model-training}

We align our LRM with human preferences using the Bradley-Terry-with-Ties (BTT) objective~\cite{liu2025improving}, an extension of BT that additionally models tie outcomes (full formulation in the appendix).
While generally effective, applying the BTT loss to our multi-margin dataset exposes a critical vulnerability.
The BTT loss continuously maximizes reward differences; consequently, for large-margin pairs (e.g., real vs. synthetic videos), the model tends to exploit superficial artifacts for trivial domain separation.
This shortcut learning causes feature collapse, degrading the sensitivity of the model to the fine-grained visual differences critical for narrow-margin pairs (e.g., intra-model pairs).
To mitigate this issue, we propose a dynamic margin-clipping mechanism guided by an Exponential Moving Average (EMA).
During training, each batch is partitioned into a narrow-margin subset $\mathcal{B}_\text{narrow}$ and a large-margin subset $\mathcal{B}_\text{large}$.
Let the predicted reward difference for a given pair be $\Delta r = R_\phi^\text{head}(F_t^\text{w}, t) - R_\phi^\text{head}(F_t^\text{l}, t)$.
We track the EMA of the reward differences exclusively on the narrow-margin subset:
\begin{equation}
    \mu \leftarrow \rho \mu + (1-\rho) \frac{1}{|\mathcal{B}_\text{narrow}|} \sum_{i \in \mathcal{B}_\text{narrow}} \Delta r_i,
\end{equation}
where $\rho=0.95$ is the decay factor.
This dynamic scalar $\mu$ serves as an adaptive threshold to bound the loss.
The reward difference for any large-margin pair $j \in \mathcal{B}_\text{large}$ is clipped by $\mu$.
The overall training objective is formulated as:
\begin{equation}
    \mathcal{L}_{\text{LRM}}(\phi) = - \mathbb{E}_{i \in \mathcal{B}_\text{narrow}} [\log \sigma(\Delta r_i)] - \mathbb{E}_{j \in \mathcal{B}_\text{large}} [\log \sigma(\min(\Delta r_j, \mu))].
\end{equation}
This adaptive clipping acts as an implicit regularizer against reward hacking.
When the reward margin of easy samples exceeds the dynamic threshold established by hard samples, their gradient contribution is zeroed out.
Consequently, the optimization is compelled to focus on fine-grained human alignment rather than exploiting superficial domain discrepancies.

\subsection{Homologous Preference Distillation: Faster and Better}
\label{homologous-preference-distillation}

\subsubsection{Key Motivation Analysis: Mitigating Multi-Objective Conflicts.}

We formulate the joint training of adversarial distillation and preference alignment on flow states $\mathbf{z}_t \in \mathcal{Z}$ as a multi-objective optimization problem.
The distillation objective $\mathcal{L}_{\text{ADV}}^G$ focuses on data distribution matching, whereas the preference objective $\mathcal{L}_\text{ReFL}$ targets specific human preference.
This task divergence naturally induces gradient conflicts.
In traditional heterogeneous paradigms, such as cross-space pixel rewards or disjoint architectures, the gradients for these two objectives are computed over unaligned representation manifolds.
Lacking a shared feature basis to constrain the optimization, the multi-objective conflict is exacerbated, often leading to out-of-distribution drift during generation.

Although our approach remains a multi-objective framework, it effectively mitigates these inherent conflicts via strict structural and data alignment.
By evaluating the identical intermediate flow state $\mathbf{z}_t$ through a shared feature backbone $F_\phi$, both tasks are anchored to a unified representation space $F_t = F_\phi(\mathbf{z}_t, t)$.
Consequently, the optimization gradients, $\nabla_{\mathbf{z}_t} \mathcal{L}_{\text{ADV}}^G$ and $\nabla_{\mathbf{z}_t} \mathcal{L}_\text{ReFL}$, are constrained by the exact same semantic basis.
This homologous formulation acts as an implicit gradient regularizer.
Instead of explicitly eliminating the objective divergence, it restricts the optimization directions within a shared manifold, promoting gradient alignment.
This mechanism ensures that the generator stably converges to the optimal joint support of visual fidelity and human preference, bypassing the need for computationally expensive explicit gradient matching.

\input{algorithm/homologous_PD}

\subsubsection{Homologous Model Design}
To actualize the homologous mechanism, we design a modular dual-head architecture.
Our LRM serves as a shared backbone for latent feature extraction, branching into two parallel heads to support simultaneous preference alignment and adversarial distillation: a frozen reward head ${R}_\phi^\text{head}(\cdot,t)$ and a trainable discriminator head ${D}_\psi^\text{head}(\cdot,t)$.
To ensure structural simplicity, we initialize the discriminator ${D}_\psi^\text{head}$ directly from the weights of the reward head ${R}_\phi^\text{head}$.
This direct weight inheritance provides a robust semantic inductive bias.
By transferring the comprehensive visual understanding of the reward model to the discriminator, this unified design stabilizes the early phase of adversarial training and accelerates convergence.

\paragraph{Unified Feature Routing.}
As shown in Fig.~\hyperref[fig:main_2]{\ref*{fig:main_2}-c}, the forward pass operates as a synchronized evaluation pipeline.
The generator $v_\theta$ first predicts a velocity vector to reconstruct the clean latent representation $\hat{\mathbf{z}}_0$.
Because the shared backbone is pre-trained on intermediate flow states, we construct the corresponding flow state via linear interpolation: $\hat{\mathbf{z}}_t = (1-t)\hat{\mathbf{z}}_0 + t \mathbf{z}_1$, where $\mathbf{z}_1 \sim \mathcal{N}(0; \mathbf{I})$.
The shared backbone then extracts the latent features $F_t = F_\phi(\hat{\mathbf{z}}_t, t)$.
We route these features in parallel to compute the preference score $r = R_\phi^\text{head}(F_t, t)$ and the adversarial logit $d = D_\psi^\text{head}(F_t, t)$.
Evaluating both tasks on identical features strictly enforces representational alignment, ensuring that the gradient updates for distillation and preference are anchored to the exact same semantic domain.

\subsubsection{Policy Distillation Objective.}

We optimize the generator $v_\theta$ and a trainable discriminator head ${D}_\psi^\text{head}$, while keeping the shared backbone $F_\phi$ and the reward head ${R}_\phi^\text{head}$ frozen.
The discriminator head is updated via the Hinge adversarial loss, denoted as $\mathcal{L}_{\text{ADV}}^D$.
Concurrently, the generator minimizes the adversarial loss $\mathcal{L}_{\text{ADV}}^G$ to align with the real data distribution and maximizes the preference score $\mathcal{L}_{\text{ReFL}}(\theta) = -\mathbb{E}[R_\phi^\text{head}(F_t, t)]$ provided by the frozen reward head.
However, a direct linear combination of these objectives causes severe optimization instability.
During the early training phase, the generated states deviate from the target data distribution.
Enforcing alignment thus produces unreliable gradients.

\paragraph{Adaptive Preference Weight.}
To ensure stable convergence, we introduce an adaptive preference weighting mechanism.
The discriminator loss is high for easily distinguishable artifacts,
and approaches a Nash equilibrium value of $0.5$ for realistic samples.
We project the deviation from this equilibrium into a target preference weight $\lambda$ using a half-Gaussian squash function:
\begin{equation}
\hat{\lambda} = \exp\left( -\text{ReLU}\left(\text{sg} \left( \mathcal{L}_\text{ADV}^{\text{D}} \right) - 0.5\right)^2 \right).
\end{equation}
This continuous mapping operates as a soft bounding constraint.
When the generated state diverges from the target data distribution ($\mathcal{L}_\text{ADV}^{\text{D}} \rightarrow \infty$), the exponential function heavily penalizes the target weight ($\hat{\lambda} \rightarrow 0$).
This step mathematically suppresses the preference gradient.
As structural realism improves, the loss approaches the equilibrium ($\mathcal{L}_\text{ADV}^{\text{D}} \rightarrow 0.5$).
The penalty vanishes, smoothly activating the preference objective ($\hat{\lambda} \rightarrow 1$).
To defend against erratic spikes in the discriminator loss, we apply an exponential moving average.
This operation decouples the dynamic weight from single-step fluctuations:
\begin{equation}
\lambda = \beta \cdot \lambda^{-} + (1 - \beta) \cdot \hat{\lambda},
\end{equation}
where $\beta = 0.99$ acts as the momentum factor.
This dual constraint mathematically guarantees the safe and gradual introduction of the preference gradient.
The preference gradient activates only when the generator exhibits sustained realism.
The total training objective for the generator is formulated as:
\begin{equation}
\mathcal{L}_{\text{HPD}}(\theta) = \mathcal{L}_{\text{ADV}}^G + \lambda \cdot \mathcal{L}_{\text{ReFL}}.
\end{equation}
By dynamically coupling these objectives, this formulation acts as a self-paced curriculum.
It safely anchors the optimization trajectory and ensures the stable convergence of the generator into the optimal joint distributional support.

%% file: figures/main_2.tex
\begin{figure}[t]
  \centering
  \includegraphics[width=0.95\textwidth]{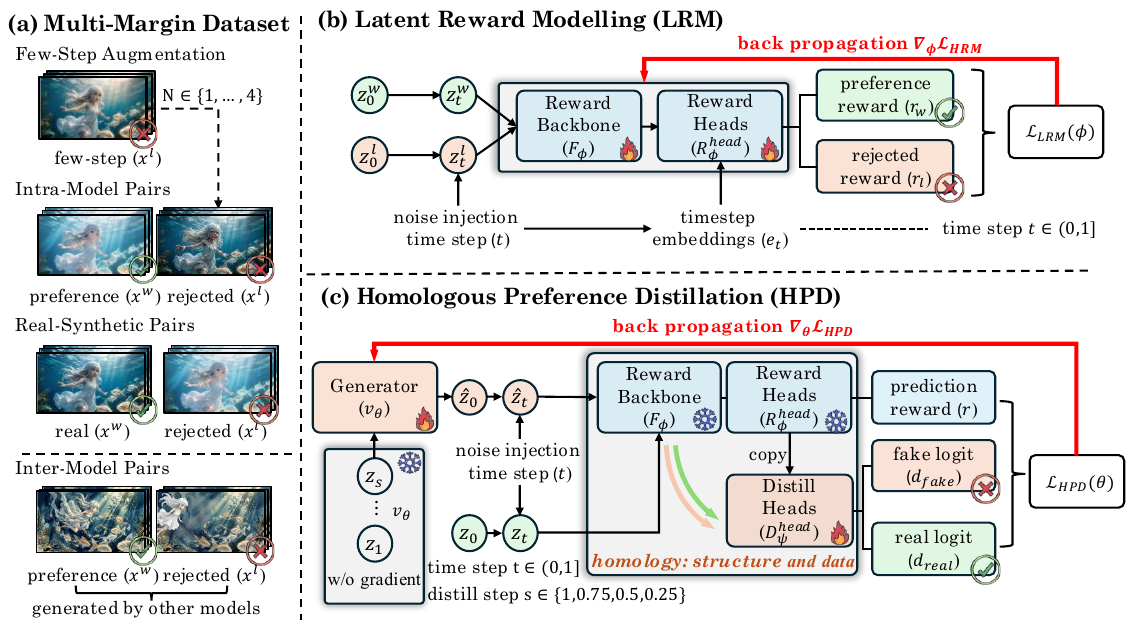}  
  \caption{
  \textbf{Overall architecture of Reward Lightning.}
  (a) Dataset: Fuzzy videos generated by $1-4$ NFEs as rejected videos to adapt to subsequent preference distillation.
  Other video pairs are used to improve the generalization of LRM.
  (b) LRM: Evaluates video pairs in latent space; the reward backbone inherits from a pre-trained model, while the reward head outputs video scores.
  We employ the Bradley-Terry with Ties (BTT) loss and introduce a regularization loss to prevent reward hacking.
  (c) HPD: The distilled head is extended from a reward head with identical architecture. 
  Both heads interact with the same reward features in parallel, outputting rewards and logits. 
  We provide dynamic preference weighting to ensure stability during initial training.
  }
  \label{fig:main_2}
\end{figure}

%% file: algorithm/homologous_PD.tex
\begin{algorithm}[t]
    \caption{Homologous Preference Distillation}
    \label{alg:hpd}
    \resizebox{0.95\linewidth}{!}{%
    \begin{minipage}{\linewidth}
    \begin{algorithmic}[1]
        \REQUIRE Generator $v_\theta$, Reward backbone $F_\phi$, Reward head $R_\phi^\text{head}$, Dataset $\mathcal{D}_\text{train}$
        \STATE \textbf{Initialize:} Discriminator $D_\psi^\text{head}$ weights $\leftarrow R_\phi^\text{head}$
        \WHILE{not converged}
            \STATE {\color{gray}\textbf{// Distillation Sampling}}
            \STATE Sample $\mathbf{z}_{1} \sim \mathcal{N}(0; \mathbf{I})$, $\{\textbf{z}_0, c\} \sim \mathcal{D}_\text{train}$ and $s\in\{1,0.75,0.5,0.25\}$
            \STATE Compute: $\mathbf{z}_s = \text{sg}\left( \text{ODESolver}(v_\theta, \mathbf{z}_{1}, c, 1 \rightarrow s) \right)$
            \STATE Compute: $\hat{\mathbf{z}}_0 = \mathbf{z}_s - s \cdot v_\theta(\mathbf{z}_s, c)$
            \STATE {\color{gray}\textbf{// Construct Homologous States}}
            \STATE Sample $\textbf{z}_1 \sim \mathcal{N}(0; \mathbf{I})$ and $t \sim \mathcal{U}[0, 1]$
            \STATE Compute: $\hat{\mathbf{z}}_t = (1-t)\hat{\mathbf{z}}_0 + t \textbf{z}_1; \quad \mathbf{z}_t = (1-t)\mathbf{z}_0 + t \textbf{z}_1$
            \STATE {\color{gray}\textbf{// Discriminator Optimization}}
            \STATE Compute: $F_t^\text{fake} = F_\phi(\hat{\mathbf{z}}_t, t); \quad F_t^\text{real} = F_\phi(\mathbf{z}_t, t)$
            \STATE Compute: $\mathcal{L}_\text{ADV}^D = \frac{1}{2}(\max(0,1-D_\psi^\text{head}(F_\text{real}, t)) + \max(0, 1+D_\psi^\text{head}(F_\text{fake}, t)))$
            \STATE Update: $\psi \leftarrow \psi - \eta_D \nabla_\psi \mathcal{L}_\text{ADV}^D$
            \STATE {\color{gray}\textbf{// Generator Optimization: Adversarial and ReFL loss}}
            \STATE Compute: $\mathcal{L}_\text{ReFL} = -R_\phi^\text{head}(F_\text{fake}, t)$
            \STATE Compute: $\mathcal{L}_\text{ADV}^G = -D_\psi^\text{head}(F_\text{fake}, t)$
            \STATE Compute: $\lambda = \exp\left( -\text{ReLU}\left(\text{sg} \left( \mathcal{L}_\text{ADV}^D \right) - 0.5\right)^2 \right)$
            \STATE Compute: $\mathcal{L}_\text{HPD} = \mathcal{L}_\text{ADV}^G - \lambda \cdot \mathcal{L}_\text{ReFL}$
            \STATE Update: $\theta \leftarrow \theta - \eta_G \nabla_\theta \mathcal{L}_\text{HPD}$
        \ENDWHILE
    \end{algorithmic}
    \end{minipage}}
\end{algorithm}

%% file: 04_experiment.tex
\input{tables/main_1}
\input{tables/main_2}

\input{figures/main_4}
\input{figures/main_5}

\section{Experiments}

\subsection{LRM: Reward Learning}
\label{lrm:reward-learning}

\subsubsection{Datasets \& Training Settings.}
\textit{Datasets.} 
We train the LRM using the multi-margin dataset described in \cref{multi-margin-dataset-collection}.
All captions and prompts are from Koala-36M~\cite{Wang_2025_CVPR}.
We provide comprehensive statistics and human annotation details in the appendix.
\textit{Training Settings.}
We employ Wan2.2-14B~\cite{wan2025wan} as the base architecture for all reward modeling experiments.
The reward model is optimized using AdamW~\cite{loshchilov2018decoupled} with betas of $0.9$ and $0.999$, a weight decay of $1\times10^{-3}$, and a constant learning rate of $1\times10^{-5}$.
We process the 5-second paired videos at a resolution of 720P.
For memory management, we apply three memory optimization strategies.
(1) Offload: we transfer frozen models (e.g., VAE~\cite{2014diederikAuto} and text encoder~\cite{raffel2020t5}) between CPUs and GPUs, which reduces the peak memory before backpropagation.
(2) Fully Sharded Data Parallel (FSDP)~\cite{zhao2023pytorchfsdpexperiencesscaling}: sharding model parameters, gradients, and optimizer states across GPUs;
(3) Ulysses Sequence Parallelism (SP)~\cite{jacobs2023deepspeedulyssesoptimizationsenabling}: partitioning latent video tokens within self-attention blocks to enable efficient processing of long temporal sequences.

\subsubsection{Benchmarks \& Evaluation Metrics.}
To evaluate the human alignment and generalization of the reward model, we employ two popular out-of-distribution (OOD) video reward benchmarks.
(1) VideoGen-RewardBench~\cite{liu2025improving}: This benchmark generates video pairs based on prompts from VideoGen-Eval, which includes different modern T2V models.
(2) GenAI-Bench~\cite{jiang2024genai}: It generates short video pairs using early T2V models, providing a low-quality distribution for evaluating generalization.
Given the pairwise preference labels, we adopt preference accuracy as our evaluation metric, which directly reflects the agreement between model predictions and human annotations.
We compare LRM against baselines including pixel reward models~\cite{he2024videoscore,xu2024visionreward,liu2025improving,wang2024lift} and latent ones~\cite{mi2025video}.

\subsubsection{Main Results.}
\cref{main_1} demonstrates that our LRM achieves state-of-the-art preference accuracy on VideoGen-RewardBench~\cite{liu2025improving} and GenAI-Bench~\cite{jiang2024genai}, reaching 72.24\% and 59.85\%, respectively, surpassing VideoAlign~\cite{liu2025improving}, the strongest pixel-level baseline, by over 10\%. 
These results reflect precise alignment with human judgment across three core dimensions: Text Alignment, comprising subject and background consistency;
Motion Quality, involving smoothness and dynamic degree; and Visual Quality, which includes aesthetic and image quality.
LRM also improves training efficiency by operating directly in the latent space rather than the pixel space.
By avoiding expensive VAE decoding and high-dimensional pixel convolutions, LRM lowers both training memory and latency at 720P.

\subsubsection{Ablation Study.}
Tab.~\hyperref[tab:ablation]{\ref*{tab:ablation}-(a-c)} highlights the contribution of each component to LRM performance. The dataset composition study in part (a) establishes that the full multi-margin collection is indispensable for preference alignment, as excluding few-step augmentation or real-synthetic pairs leads to a significant decline in accuracy and reward scores. In terms of head architecture, time-step projection in part (b) proves superior to self-attention and trainable query baselines. This design allows the model to adaptively extract robust features from the latent space $z$ by directly conditioning on the temporal embedding across the flow trajectory. Reward regularization in part (c) is also critical for preventing reward hacking, as it mathematically bounds reward gaps for stable feedback. 

\subsection{HPD: Improving and Accelerating Diffusion Models}
\subsubsection{Datasets \& Training Settings.}
Unlike the paired video datasets, we train the generator on single captioned videos.
We collect a high-quality real-video dataset comprising Koala-36M~\cite{Wang_2025_CVPR} and Intern4K~\cite{lin2025cascadev}.
We build our generator on Wan2.2-14B~\cite{wan2025wan}, and initialize the discriminator from LRM.
In the training process, we employ AdamW~\cite{loshchilov2018decoupled} with betas of 0.5 and 0.999 for lower memory consumption.
This optimizer adopts a weight decay of $1\times10^{-3}$ and a constant learning rate of $5\times10^{-7}$.
The generator is trained with the batch size of 64 and 2,000 steps.
An EMA of 0.995 is applied to the generator.
For memory efficiency, we extend the offloading strategy to the EMA generator in addition to the frozen VAE and text encoder.
Furthermore, we employ a full-sharding strategy via FSDP and SP across the generator, reward model, and discriminator to enable stable training within the limited hardware constraints.

\subsubsection{Benchmarks \& Evaluation Metrics.}
We evaluate the generator on VBench~\cite{huang2024vbench} for comprehensive video assessment.
The assessment involves multi-dimensional metrics, specifically Visual Quality, Motion Quality, and Text Alignment.
To quantify the distribution gap between generated and real-world videos, we calculate the Fréchet Video Distance (FVD).
Additionally, we conduct a human evaluation based on voting results to capture subjective visual preferences.

\subsubsection{Baseline Models.}
We evaluate HPD against state-of-the-art baselines in three categories: heterogeneous training, standalone distillation, and preference alignment.
For heterogeneous paradigms, we include DMDR~\cite{jiang2025distribution} and FlashDMD~\cite{chen2025flash}, which represent joint optimization using disjoint representation spaces.
The distillation-only baselines comprise TurboDiffusion~\cite{zhang2025turbodiffusion} and DMD2~\cite{yin2024improved} to represent current sampling acceleration techniques.
For preference alignment, we compare against PRFL~\cite{mi2025video}, a multi-step reward feedback method.

\subsubsection{Main Results.}
\cref{main_2} summarizes the quantitative performance of HPD under the 4-NFE and 1-NFE few-step settings on both VBench (T2V) and VBench-I2V~\cite{huang2024vbench}. Compared to both pure distillation and heterogeneous training paradigms, our approach achieves a superior balance between inference efficiency and generation quality. In the 4-NFE setting, our framework attains the highest overall quality score among heterogeneous methods such as DMDR~\cite{jiang2025distribution} and FlashDMD~\cite{chen2025flash}, leading across text alignment, motion quality, and visual quality. This indicates that homologous alignment effectively mitigates the multi-objective conflicts that often degrade performance in disjoint architectures. While pure distillation baselines like TurboDiffusion~\cite{zhang2025turbodiffusion} and DMD2~\cite{yin2024improved} focus primarily on acceleration, they lack the semantic guidance provided by human preference models. Our approach addresses this by integrating preference alignment directly into the distillation trajectory, yielding higher scores across three core dimensions: Text Alignment, Motion Quality, and Visual Quality. Even at the extreme limit of 1-NFE, our approach consistently surpasses the APT baseline in text alignment and motion quality, and attains the best overall quality score. As shown in \crefrange{fig:main_4}{fig:main_5}, HPD eliminates the motion blur and temporal flickering prevalent in few-step sampling, producing stable motion and crisp visual details. The human evaluation in \cref{fig:main_6} corroborates these gains, with annotators favoring our results over competing baselines.

\input{tables/main_4}
\input{figures/main_6}

\subsubsection{Ablation Study.}
Tab.~\hyperref[tab:ablation]{\ref*{tab:ablation}-(d-e)} underscores the necessity of structural homology and adaptive weighting within HPD. Initializing the discriminator with LRM parameters in part (d) provides a consistent semantic prior that raises the quality ceiling and accelerates convergence compared to random baselines. In part (e), the adaptive weight $\lambda$ ensures training stability by delaying preference alignment until the generator achieves structural realism. This self-paced mechanism prevents early gradient interference and out-of-distribution drift, guiding the model toward the optimal joint support of visual fidelity and human preference.

\subsection{Further Analysis}
\label{further-analysis}
\subsubsection{Gradient Conflict Analysis.}
\input{tables/main_5}
To validate that homology mitigates these conflicts, we measure the cosine similarity between the adversarial gradient $\nabla_\theta \mathcal{L}_{\text{ADV}}^G$ and the preference gradient $\nabla_\theta \mathcal{L}_{\text{ReFL}}$, averaged over 1{,}000 training iterations (Tab.~\hyperref[tab:gradient_conflict]{\ref*{tab:further_analysis}-a}).
In heterogeneous paradigms with a pixel-space or separate-backbone reward model, the two gradients exhibit negative cosine similarity ($-0.14$ and $-0.06$), confirming that the optimization directions actively conflict.
By contrast, HPD evaluates both objectives through a shared backbone on identical latent representations, raising the cosine similarity to $0.41$.
This confirms that structural and space homology acts as an implicit gradient regularizer.

\subsubsection{Efficiency \& Quality Analysis.}
\input{figures/main_7}
As shown in \cref{fig:main_7}, across 1 to 8 sampling steps, HPD maintains a superior efficiency-quality profile over DMD2~\cite{yin2024improved}, TurboDiffusion~\cite{zhang2025turbodiffusion}, DMDR~\cite{jiang2025distribution}, and FlashDMD~\cite{chen2025flash}.
Unlike baselines that suffer structural drift, our homologous design anchors the trajectory within a unified manifold, delivering higher fidelity at lower cost.
Representational homology thus enables near-teacher performance at 1-NFE, closing the speed--alignment gap. Notably, this advantage widens as the step count decreases, indicating that representational consistency matters most in the aggressive few-step regime.

\subsubsection{Distillation Generalizability Analysis.}

As shown in Tab.~\hyperref[main_3]{\ref*{tab:further_analysis}-b}, LRM yields consistent gains across paradigms like rCM~\cite{zheng2025rcm} and DMD2~\cite{yin2024improved}.
However, the largest gains arise within HPD, where homology minimizes gradient conflicts and remains vital for the optimal ceiling in acceleration and human alignment.

%% file: tables/main_1.tex
\begin{table}[t]
\centering
\caption{
\textbf{Quantitative comparison of preference accuracy and online reward feedback efficiency.}
We report preference accuracy on out-of-distribution benchmarks, together with the
training latency, memory consumption, and throughput of online reward feedback.
All computations run on $8\times$ H20 GPUs.
\underline{Underline}: The second-best performance. \textbf{Bold}: Best performance.
}
\label{main_1}
\resizebox{0.95\textwidth}{!}{%
\begin{tabular}{lcccccccc}
\toprule
\multirow{3}{*}{\textbf{Method}} & \multicolumn{2}{c}{\textbf{VideoGen-RewardBench}} & \multicolumn{2}{c}{\textbf{GenAI-Bench}} & \multicolumn{4}{c}{\textbf{Reward Feedback Efficiency}} \\
\cmidrule(lr){2-3} \cmidrule(lr){4-5} \cmidrule(lr){6-9}
 & \multicolumn{2}{c}{Overall Accuracy ($\uparrow$)} & \multicolumn{2}{c}{Overall Accuracy ($\uparrow$)}
 & \multirow{2}{*}{\makecell{Resolution}}
 & \multirow{2}{*}{\makecell{Frames}}
 & \multirow{2}{*}{\makecell{Train \\ Memory}}
 & \multirow{2}{*}{\makecell{Train \\ Latency}} \\
\cmidrule(lr){2-3} \cmidrule(lr){4-5}
 & w/ Ties & w/o Ties & w/ Ties & w/o Ties & & & & \\
\midrule
\rowcolor{gray!20} \multicolumn{9}{l}{\textbf{Pixel Reward Models}} \\
\midrule
Random (QwenVL-2~\cite{wang2024qwen2})  & 41.86 & 50.30 & 33.67 & 49.84 & $480\times480$   & $\le$ 1 & -  & - \\
LiFT~\cite{wang2024lift}                & 39.08 & 57.26 & 37.06 & 58.39 & $384\times384$   & $\le$ 1 & 87.81  & 213.2 \\
VideoScore~\cite{he2024videoscore}      & 41.80 & 50.22 & 49.03 & 71.69 & $384\times 384$  & $\le$ 1 & 85.60  & 207.4 \\
VisionRewrd~\cite{xu2024visionreward}   & 56.77 & 67.59 & 51.56 & 72.41 & $224\times224$   & $\le$ 1 & 82.15  & 185.3 \\
VideoAlign~\cite{liu2025improving}       & \underline{61.26} & \underline{73.59} & 49.41 & \underline{72.89} & $448\times448$   & $\le$ 1 & 87.46  & 217.9 \\
\rowcolor{gray!20} \multicolumn{9}{l}{\textbf{Latent Reward Models}} \\
\midrule
Random (Wan2.2~\cite{wan2025wan})        & 48.41 & 58.03 & 40.12 & 53.88 & $1280 \times 720$ & $\ge$ 81 & - & - \\
PRFL~\cite{mi2025video}                 & 57.59 & 69.20 & \underline{53.40} & 72.10 & $1280 \times 720$ & $\ge$ 81 & 77.32 & 156.8 \\
\rowcolor{bestblue} \textbf{Ours}  & \textbf{72.24} & \textbf{86.63} & \textbf{59.85} & \textbf{78.44} & $1280 \times 720$ & $\ge$ 81 & 76.24 & 154.2 \\
\bottomrule
\end{tabular}%
}
\end{table}

%% file: tables/main_2.tex
\begin{table}[t]
\centering
\caption{
\textbf{Quantitative comparison of Text-to-Video (T2V) and Image-to-Video (I2V) on VBench.}
For a fair comparison, we build all models on the Wan2.2 architecture. 
With the exception of TurboDiffusion, we reproduce all results ourselves.
VQ, MQ, and TA denote Visual Quality, Motion Quality, and Text Alignment.
}
\label{main_2}
\resizebox{0.95\textwidth}{!}{%
\begin{tabular}{lc cccc cccc}
\toprule
& & \multicolumn{4}{c}{\textbf{VBench}} & \multicolumn{4}{c}{\textbf{VBench-I2V}} \\
\cmidrule(lr){3-6} \cmidrule(lr){7-10}
\multirow{-2}{*}{\textbf{Methods}} & \multirow{-2}{*}{\makecell{\textbf{Training} \textbf{Type}}} & \makecell{\textbf{TA} ($\uparrow$)} & \makecell{\textbf{MQ} ($\uparrow$)} & \makecell{\textbf{VQ} ($\uparrow$)} & \makecell{\textbf{Quality} \\ \textbf{Score} ($\uparrow$)} & \makecell{\textbf{TA} ($\uparrow$)} & \makecell{\textbf{MQ} ($\uparrow$)} & \makecell{\textbf{VQ} ($\uparrow$)} & \makecell{\textbf{I2V} \\ \textbf{Score} ($\uparrow$)} \\
\midrule
\rowcolor{gray!20} \multicolumn{10}{l}{\textbf{80-NFEs}} \\
\midrule
Wan2.2~\cite{wan2025wan} & Pretrain & 97.34 & 79.59 & 69.48 & 84.23 & 95.87 & 74.64 & 67.63 & 92.91 \\
RGB ReFL~\cite{lin2025contentv} & PixelRL & 97.45 & 80.12 & 69.55 & 84.41 & 96.12 & 75.24 & 67.85 & 93.52 \\
PRFL~\cite{mi2025video} & LatentRL & 97.83 & 81.56 & 70.45 & 85.72 & 96.65 & 76.58 & 68.61 & 94.87 \\
\rowcolor{gray!20} \multicolumn{10}{l}{\textbf{4-NFEs}} \\
\midrule
Wan2.2~\cite{wan2025wan} & Pretrain & 80.15 & 61.24 & 56.43 & 68.52 & 80.41 & 61.81 & 55.94 & 75.97 \\
TurboDiffusion~\cite{zhang2025turbodiffusion} & Distill & 94.53 & \underline{75.62} & 63.28 & 83.51 & 95.84 & \underline{74.25} & 64.87 & 92.45 \\
DMD2~\cite{yin2024improved} & Distill & 91.46 & 72.53 & 64.38 & 81.27 & 94.28 & 73.33 & 67.78 & 91.92 \\
DMDR~\cite{jiang2025distribution} & PixelRL \& Distill & \underline{96.14} & 71.85 & \underline{68.57} & \underline{84.63} & \underline{96.17} & 72.14 & \underline{69.56} & \underline{93.28} \\
FlashDMD~\cite{chen2025flash} & PixelRL \& Distill & 88.19 & 68.23 & 67.84 & 82.36 & 92.15 & 69.45 & 69.12 & 92.59 \\
\rowcolor{bestblue} \textbf{HPD} & LatentRL \& Distill & \textbf{97.26} & \textbf{77.48} & \textbf{68.71} & \textbf{85.83} & \textbf{96.32} & \textbf{75.39} & \textbf{69.78} & \textbf{95.34} \\
\rowcolor{gray!20} \multicolumn{10}{l}{\textbf{1-NFEs}} \\
\midrule
Wan2.2~\cite{wan2025wan} & Pretrain & 70.45 & 53.27 & 51.18 & 61.26 & 72.84 & 55.73 & 51.89 & 69.59 \\
APT~\cite{lin2025diffusion} & Distill & \underline{85.37} & \underline{65.42} & \textbf{64.19} & \underline{78.53} & \underline{87.89} & \underline{68.18} & \underline{61.81} & \underline{84.87} \\
\rowcolor{bestblue} \textbf{HPD} & LatentRL \& Distill & \textbf{92.54} & \textbf{70.16} & \underline{62.83} & \textbf{82.42} & \textbf{95.14} & \textbf{73.39} & \textbf{65.31} & \textbf{91.92} \\
\bottomrule
\end{tabular}%
}
\end{table}

%% file: figures/main_4.tex
\begin{figure}[tb]
  \centering
  \includegraphics[width=0.95\textwidth]{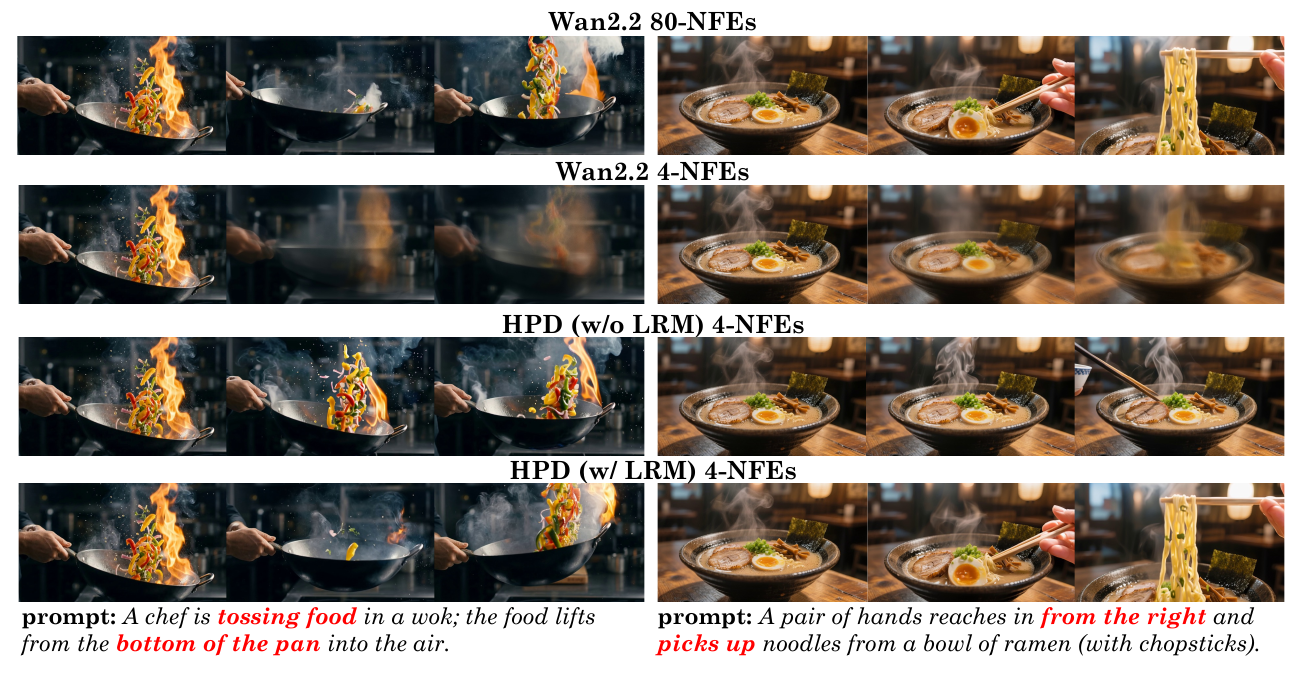}  
  \caption{
  \textbf{Qualitative Results.}
  Upper: Visualization of $80$-NFEs and $4$-NFEs from the Wan2.2-I2V-A14B.
  Bottom: Built on the baseline model, compared against a distillation without homologous preference distillation.
  }
  \label{fig:main_4}
\end{figure}

%% file: figures/main_5.tex
\begin{figure}[htb]
  \centering
  \includegraphics[width=0.95\textwidth]{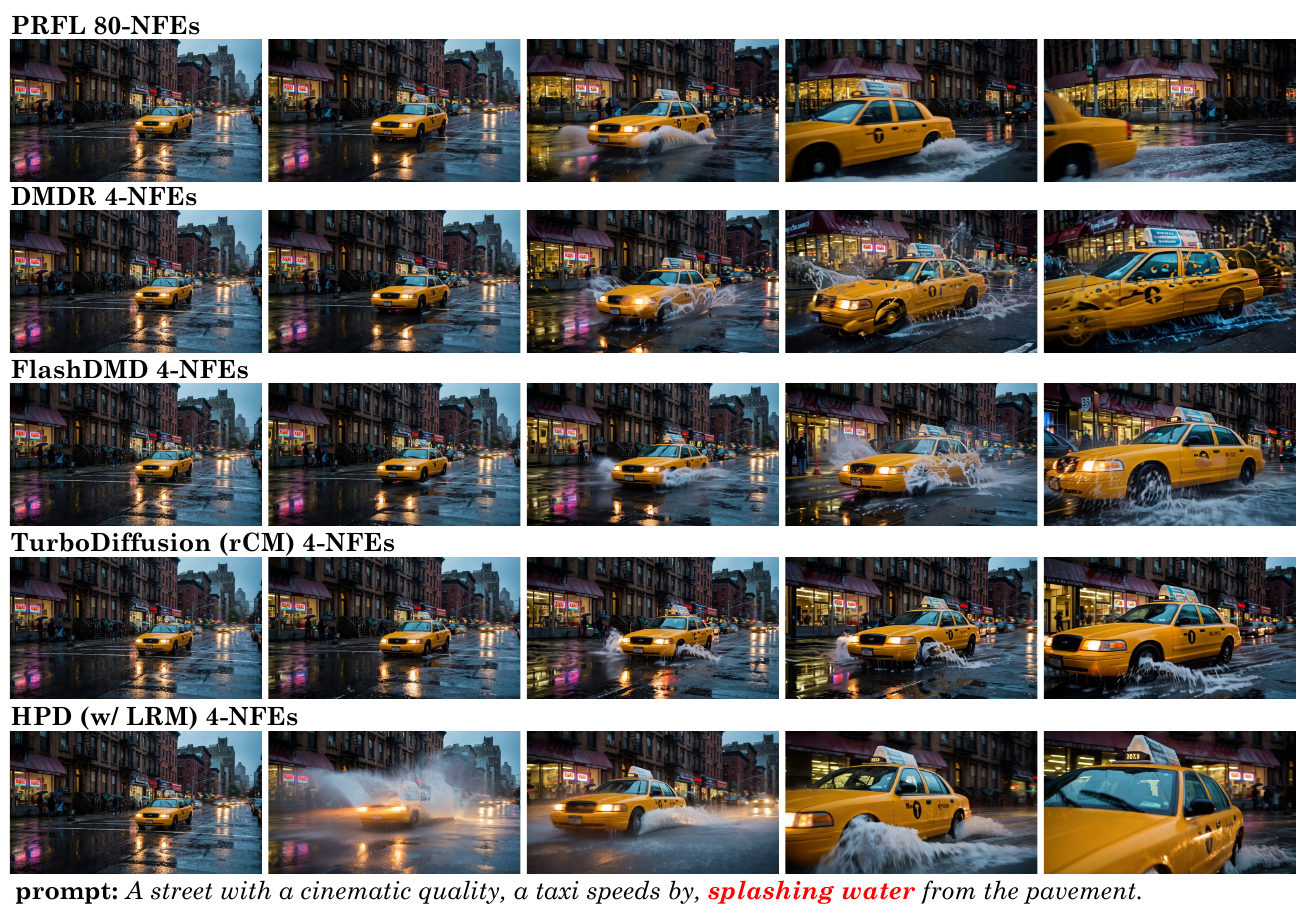}  
  \caption{
  \textbf{Qualitative Results}
    For fair comparison, all models are built on Wan2.2-I2V-A14B.
    Except for TurboDiffusion, we reproduce all results ourselves.
    \textit{Row} $1$: ReFL model based on latent reward model.
    \textit{Row} $2-3$: Heterogeneous model, composed of pixel ReFL and DMD distillation.
    \textit{Row} $4$: Distillation model, primarily based on rCM method.
    \textit{Row} $5$: Homologous model, with structural and space homology.
  }
  \label{fig:main_5}
\end{figure}

%% file: tables/main_4.tex
\begin{table}[t]
\centering
\scriptsize
\setlength{\tabcolsep}{4pt}
\captionsetup{position=top,skip=4pt}
\captionsetup[sub]{position=top,skip=2pt}
\caption{
\textbf{Ablation study on key components.} (a)-(c) LRM: We provide ablations on the dataset composition, head architecture, and reward loss functions. (d)-(e) HPD: We ablate the distilled parameters and preference-loss weighting.
}
\label{tab:ablation}
\subcaptionbox{\textbf{Dataset Composition}\label{ab:a}}[0.334\textwidth]{%
\resizebox{\linewidth}{!}{%
\begin{tabular}{@{}lcc@{}}
\toprule
\textbf{Setting} & \textbf{Acc.}\,($\uparrow$) & \textbf{Score}\,($\uparrow$) \\
\midrule
\rowcolor{gray!20} Full & 72.24 & 95.34 \\
\textit{w/o} Few-Step & 68.47 & 93.81 \\
\textit{w/o} Real-Syn. & 65.12 & 92.45 \\
\bottomrule
\end{tabular}}}
\hfill
\subcaptionbox{\textbf{Head Architecture}\label{ab:b}}[0.337\textwidth]{%
\resizebox{\linewidth}{!}{%
\begin{tabular}{@{}lcc@{}}
\toprule
\textbf{Setting} & \textbf{Acc.}\,($\uparrow$) & \textbf{Score}\,($\uparrow$) \\
\midrule
Self-Attention & 64.38 & 91.87 \\
\rowcolor{gray!20} \textit{w/} Query & 69.85 & 94.12 \\
\textit{w/} Time Proj. & 72.24 & 95.34 \\
\bottomrule
\end{tabular}}}
\hfill
\subcaptionbox{\textbf{Reward Model Loss}\label{ab:c}}[0.280\textwidth]{%
\resizebox{\linewidth}{!}{%
\begin{tabular}{@{}lcc@{}}
\toprule
\textbf{Setting} & \textbf{Acc.}\,($\uparrow$) & \textbf{Score}\,($\uparrow$) \\
\midrule
\textit{w/o} reg. & 62.15 & 90.76 \\
$\rho=0.5$ & 68.91 & 93.58 \\
\rowcolor{gray!20} $\rho=0.95$ & 72.24 & 95.34 \\
\bottomrule
\end{tabular}}}

\subcaptionbox{\textbf{Head Parameters}\label{ab:d}}[0.469\textwidth]{%
\resizebox{\linewidth}{!}{%
\begin{tabular}{@{}lcccc@{}}
\toprule
\textbf{Setting} & \textbf{TA}\,($\uparrow$) & \textbf{MQ}\,($\uparrow$) & \textbf{VQ}\,($\uparrow$) & \textbf{Score}\,($\uparrow$) \\
\midrule
Random & 91.14 & 68.58 & 62.87 & 88.26 \\
Pre-trained & 94.57 & 73.12 & 66.74 & 93.48 \\
\rowcolor{gray!20} LRM & 96.32 & 75.39 & 69.78 & 95.34 \\
\bottomrule
\end{tabular}}}
\hfill
\subcaptionbox{\textbf{Adaptive Preference Weight}\label{ab:e}}[0.476\textwidth]{%
\resizebox{\linewidth}{!}{%
\begin{tabular}{@{}lcccc@{}}
\toprule
\textbf{Setting} & \textbf{TA}\,($\uparrow$) & \textbf{MQ}\,($\uparrow$) & \textbf{VQ}\,($\uparrow$) & \textbf{Score}\,($\uparrow$) \\
\midrule
$\lambda=0$ & 91.45 & 71.34 & 67.12 & 90.87 \\
$\lambda=1$ & 93.12 & 72.18 & 67.58 & 92.73 \\
\rowcolor{gray!20} $\lambda \propto - \mathcal{L}_\text{ADV}^D$ & 96.32 & 75.39 & 69.78 & 95.34 \\
\bottomrule
\end{tabular}}}

\end{table}

%% file: figures/main_6.tex
\begin{figure}[t]
  \centering
  \includegraphics[width=0.95\textwidth]{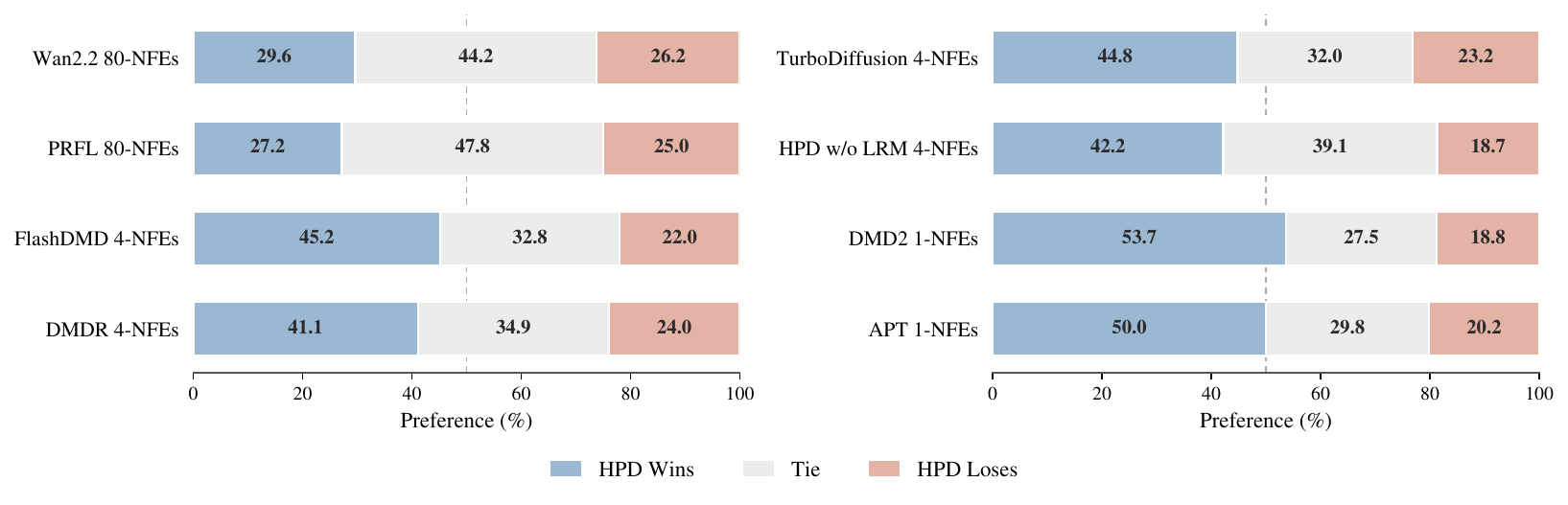}  
  \caption{
  \textbf{Human Evaluation.}
  We report the Win/Tie/Lose rate for HPD $1,4-$NFEs, compared with the Euler baseline, distillation, and heterogeneous models.
  }
  \label{fig:main_6}
\end{figure}

%% file: tables/main_5.tex
\begin{table}[t]
\centering
\scriptsize
\setlength{\tabcolsep}{4pt}
\captionsetup{position=top,skip=4pt}
\captionsetup[sub]{position=top,skip=2pt}
\caption{\textbf{Further analysis on the homologous design.} \textbf{(a)} Gradient conflict analysis: cosine similarity between $\nabla_\theta \mathcal{L}_{\text{ADV}}^G$ and $\nabla_\theta \mathcal{L}_{\text{ReFL}}$, averaged over $1{,}000$ iterations (4-NFE, I2V); positive values indicate aligned gradients. \textbf{(b)} LRM robustness and the impact of HPD homology across distillation paradigms.}
\label{tab:further_analysis}
\subcaptionbox{\textbf{Gradient Conflict}\label{tab:gradient_conflict}}[0.362\textwidth]{%
\renewcommand{\arraystretch}{2.1}%
\resizebox{\linewidth}{!}{%
\begin{tabular}{@{}lcc@{}}
\toprule
\textbf{Setting} & \textbf{Backbone} & \makecell{\textbf{Cosine}\\\textbf{Similarity}\,($\uparrow$)} \\
\midrule
\makecell[l]{Pixel Reward\\+ Distillation} & Separate & $-0.14$ \\
\midrule
\makecell[l]{Latent Reward\\+ Distillation} & Separate & $-0.06$ \\
\midrule
\textbf{HPD (Ours)} & Shared & $\mathbf{0.41}$ \\
\bottomrule
\end{tabular}}}
\hfill
\renewcommand{\arraystretch}{1}%
\subcaptionbox{\textbf{LRM Robustness}\label{main_3}}[0.616\textwidth]{%
\resizebox{\linewidth}{!}{%
\begin{tabular}{@{}lccccc@{}}
\toprule
\textbf{Methods}  & \textbf{TA} ($\uparrow$) & \textbf{MQ} ($\uparrow$) & \textbf{VQ} ($\uparrow$) & \makecell{\textbf{I2V} \\ \textbf{Score}\,($\uparrow$)} & \textbf{FVD} ($\downarrow$) \\
\midrule
rCM~\cite{zheng2025rcm}  & 95.84 & 74.25 & 64.87 & 93.45 & 207.32 \\
$+$ LRM  & 96.65 {\tiny \textcolor{blue}{+0.81}} & 74.98 {\tiny \textcolor{blue}{+0.73}} & 65.82 {\tiny \textcolor{blue}{+0.95}} & 94.32 {\tiny \textcolor{blue}{+0.87}} & 205.45 {\tiny \textcolor{blue}{-1.87}} \\
\midrule
DMD2~\cite{yin2024improved} & 94.28 & 73.33 & 67.78 & 91.92 & 218.78 \\
$+$ LRM  & 95.07 {\tiny \textcolor{blue}{+0.79}} & 74.31 {\tiny \textcolor{blue}{+0.98}} & 68.63 {\tiny \textcolor{blue}{+0.85}} & 92.71 {\tiny \textcolor{blue}{+0.79}} & 217.64 {\tiny \textcolor{blue}{-1.14}} \\
\midrule
HPD  & 94.14 & 72.92 & 67.44 & 93.09 & 167.14 \\
$+$ LRM  & 96.32 {\tiny \textcolor{blue}{+2.18}} & 75.39 {\tiny \textcolor{blue}{+2.47}} & 69.78 {\tiny \textcolor{blue}{+2.34}} & 95.34 {\tiny \textcolor{blue}{+2.25}} & 162.87 {\tiny \textcolor{blue}{-4.27}} \\
\bottomrule
\end{tabular}}}
\end{table}

%% file: figures/main_7.tex
\begin{wrapfigure}[17]{r}{0.46\textwidth}
  \centering
  \includegraphics[width=0.40\textwidth]{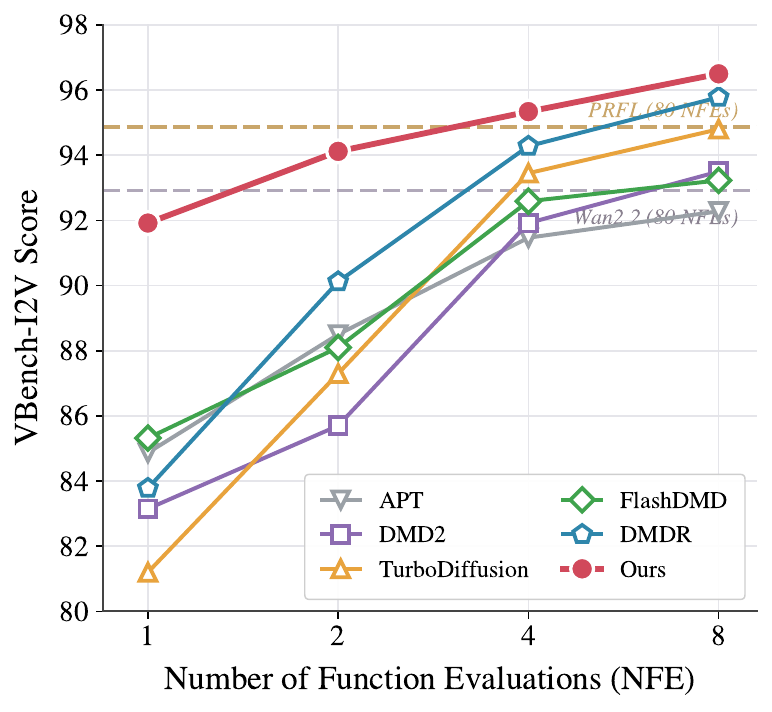}
  \caption{
  \textbf{Efficiency-quality Pareto front on Wan2.2-I2V-14B.} Dotted lines denote 80-NFE baselines. HPD dominates the top-left region across 1 to 8 NFEs.
  }
  \label{fig:main_7}
\end{wrapfigure}

%% file: 06_conclusion.tex
\section{Conclusion}

We have introduced Reward Lightning, a unified framework that aligns representations through structural and space homology for joint preference alignment and distillation acceleration.
Evaluating both objectives within an identical manifold minimizes the angle between conflicting gradients, acting as an implicit regularizer that drives the generator toward the joint support of high fidelity and human alignment.
It accelerates inference well beyond heterogeneous methods.

%% file: 07_appendix.tex
\begin{center}
    \Large
\textbf{Reward Lightning: Fast Video Generation via Homologous Preference Distillation}\\
    \Large
    Supplementary Material\\
\end{center}

The supplementary material is organized as follows:
\begin{itemize}
    \item Appendix~\ref{mathematical-theory}: The mathematical theory of background knowledge. 
    \item Appendix~\ref{more-annotation-details}: The annotation protocols for preference datasets. 
    \item Appendix~\ref{more-implementation-details}: Extensive training implementation details for baseline methods. 
    \item Appendix~\ref{more-experimental-results}: Additional experimental results. 
    \item Appendix~\ref{limitation-and-future-works}: Discusses our limitations and directions for future work. 
\end{itemize}

\section{Mathematical Theory}
\label{mathematical-theory}
\subsection{Rectified Flow}
Rectified flow~\cite{liu2023flow,lipman2023flow} learns an ordinary differential equation (ODE)~\cite{song2021scorebased} that deterministically maps a Gaussian prior to a complex target distribution.
This process operates within a continuous latent space.
Let $\mathbf{z}_1 \sim \mathcal{N}(\mathbf{0}, \mathbf{I})$ be a Gaussian noise vector and $\mathbf{z}_0 \sim p_\text{data}$ a target latent sample.
Rectified flow interpolates the two along a straight-line path for $t \in (0, 1]$:
\begin{equation}
    \mathbf{z}_t = t \mathbf{z}_1 + (1 - t) \mathbf{z}_0.
\end{equation}
The time derivative of this path is defined as:
\begin{equation}
v(\mathbf{z}_t, t) = \frac{d\mathbf{z}_t}{dt} = \mathbf{z}_1 - \mathbf{z}_0
\end{equation}
Our objective is to train a neural network $v_\theta(\mathbf{z}_t, t)$ that approximates this ideal velocity field.
We minimize the expected Mean Squared Error (MSE) over all time steps $t$:
\begin{equation}    
\begin{aligned}
\mathcal{L}_{\text{RF}}(\theta) &= \frac{1}{2} \int_0^1 \mathbb{E}_{\mathbf{z}_0 \sim p_0, \mathbf{z}_1 \sim p_1} \left[ \left\| v_\theta(\mathbf{z}_t, t) - v(\mathbf{z}_t, t) \right\|_2^2 \right] dt \\
&= \frac{1}{2} \int_0^1 \mathbb{E}_{\mathbf{z}_0, \mathbf{z}_1} \left[ \left\| v_\theta(t \mathbf{z}_1 + (1 - t) \mathbf{z}_0, t) - (\mathbf{z}_1 - \mathbf{z}_0) \right\|_2^2 \right] dt
\end{aligned}
\end{equation}
During inference, given a conditioning text prompt $c$, we start from the noise $\mathbf{z}_1$ and generate the latent variable $\mathbf{z}_0$ by solving the empirical ODE $d\mathbf{z}_t = v_\theta(\mathbf{z}_t, c, t) dt$ from $t=1$ to $t=0$.

\subsection{Bradley-Terry and Bradley-Terry with Ties}
\subsubsection{Bradley-Terry (BT) Loss.}
The standard Bradley-Terry model~\cite{terry1925bt} evaluates strict pairwise comparisons.
Given a text prompt condition $c$ and a preference dataset comprising preferred videos $\mathbf{z}_w$ and rejected videos $\mathbf{z}_l$, the model assigns scalar rewards $r_w = R_\phi(\mathbf{z}_w, c)$ and $r_l = R_\phi(\mathbf{z}_l, c)$.
We model the probability that human annotators prefer $\mathbf{z}_w$ over $\mathbf{z}_l$ as:
\begin{equation}
P(\mathbf{z}_w \succ \mathbf{z}_l) = \frac{\exp(r_w)}{\exp(r_w) + \exp(r_l)}.
\end{equation}
To align the reward model with these strict preferences, we minimize the negative log-likelihood of the observed pairs, yielding the standard BT loss.
While the BT model handles clear chosen/rejected pairs well, it inherently struggles with ties, often assigning sizeable reward margins ($\Delta r$) to tie pairs and conflating them with decisive preferences.
The training objective is defined as:
\begin{equation}    
\begin{aligned}
\mathcal{L}_{\text{BT}}(\phi) &= - \mathbb{E}_{(\mathbf{z}_w, \mathbf{z}_l, c) \sim \mathcal{D}} \left[ \log P(\mathbf{z}_w \succ \mathbf{z}_l | c) \right] \\
&= - \mathbb{E}_{(\mathbf{z}_w, \mathbf{z}_l, c) \sim \mathcal{D}} \left[ \log \left( \frac{1}{1 + \exp(-(R_\phi(\mathbf{z}_w, c) - R_\phi(\mathbf{z}_l, c)))} \right) \right] \\
&= - \mathbb{E}_{(\mathbf{z}_w, \mathbf{z}_l, c) \sim \mathcal{D}} \left[ \log \sigma \left( R_\phi(\mathbf{z}_w, c) - R_\phi(\mathbf{z}_l, c) \right) \right],
\end{aligned}
\end{equation}

\subsubsection{Bradley-Terry with Ties (BTT) Loss.}
We introduce the BTT loss~\cite{liu2025improving}, which explicitly models the probability of a tie outcome.
It introduces a control parameter $k > 1$ that governs the tendency toward ties; a larger $k$ increases the expected tie probability.
Let $y \in \{\succ, \prec, \approx\}$ denote the ground-truth preference choice (A preferred, B preferred, or Tie).
The probability distribution $P_k(y | c, \mathbf{z}_w, \mathbf{z}_l)$ over these three outcomes is defined as:
\begin{equation}
P_k(y | c, \mathbf{z}_w, \mathbf{z}_l) =
\begin{cases} 
\frac{(k^2 - 1) \exp(r_w) \exp(r_l)}{(\exp(r_w) + k \exp(r_l))(k \exp(r_w) + \exp(r_l))} & \text{Tie } (\mathbf{z}_w \approx \mathbf{z}_l), \\
\frac{\exp(r_w)}{\exp(r_w) + k \exp(r_l)} & \text{A preferred } (\mathbf{z}_w \succ \mathbf{z}_l), \\
\frac{\exp(r_l)}{k \exp(r_w) + \exp(r_l)} & \text{B preferred } (\mathbf{z}_l \succ \mathbf{z}_w).
\end{cases}
\end{equation}
We set the threshold parameter to $k = 5.0$, and train our reward model by minimizing the negative log-likelihood across all preference scenarios:
\begin{equation}
    \mathcal{L}_{BTT}(\phi) = -\mathbb{E}_{(c, \mathbf{z}_w, \mathbf{z}_l) \sim \mathcal{D}} \left[ \sum_{y \in \{\succ, \prec, \approx\}} \mathbb{1}(y) \log P_k(y | c, \mathbf{z}_w, \mathbf{z}_l) \right], \label{eq:btt}
\end{equation}
where $\mathbb{1}(y)$ is the indicator function for the observed human choice.
By optimizing the BTT objective, the reward model learns a flexible decision boundary that clusters the reward differences ($\Delta r$) of tied pairs near zero, while preserving large margins for decisive wins and losses.
This mechanism provides reliable, calibrated reward feedback for the downstream RLHF process.

\section{More Annotation Details}
\label{more-annotation-details}
\subsection{Preference Dataset}
To train a robust and generalizable Reward Model, we construct a preference dataset comprising $64,390$ video pairs.
We source the text prompts or first images from the video dataset Koala-36M~\cite{Wang_2025_CVPR} to ensure broad semantic coverage.
Given the condition $c$, we blindly present two videos generated with different random seeds.
Five professional video annotators rate each video on a 3-point Likert scale (1: Poor, 2: Acceptable, 3: Excellent) across three critical dimensions: motion dynamics, semantic alignment, and visual quality.

\subsubsection{Implementation Protocol.}
To derive definitive preference labels (win, tie, or lose) from multi-dimensional scores, we establish a deterministic evaluation protocol.
Our decision logic follows a hierarchical set of criteria:
\begin{itemize}
    \item \textit{Fundamental Flaw Filtering}: We immediately exclude any video pair exhibiting severe degradation, including black screens, subtitles, black borders, short duration, or significant visual distortion.
    \item \textit{Minimum Quality Threshold}: At least one video in the pair must score $\ge 2$ across all three dimensions. We discard pairs failing this criterion to prevent the reward model from learning suboptimal representations.
    \item \textit{Weighted Total Score}: Since motion dynamics and prompt adherence vary more significantly with random seeds than visual quality, we calculate a weighted total score using a 2:2:1 ratio for TA, MQ, and VQ, respectively.
    \item \textit{Score Comparison}: We designate the video with the higher weighted score as the winner. If the score difference is $\le 1$, we classify the pair as a tie.
\end{itemize}
For specific comparison types, we assign winners based on structural priors: real videos are preferred over synthetic ones, and multi-step over few-step samples.

\subsubsection{Quality Control and Reliable IAA Calculation.}
To ensure the reported Fleiss' Kappa ($\kappa$) is a statistically reliable metric, we implement a multi-round calibration protocol.
Before large-scale annotation, annotators independently rate a pilot batch of 200 pairs, followed by a conflict-resolution discussion to unify their subjective standards.
This repeats until the Inter-Annotator Agreement (IAA) stabilizes.
During the formal annotation phase, we rigorously compute the final Fleiss' Kappa on a randomly sampled, held-out quality assurance subset of 1,000 video pairs.
The calculation follows:
\begin{equation}
\kappa = \frac{\bar{P} - \bar{P}_e}{1 - \bar{P}_e},
\end{equation}
where $\bar{P}$ represents the observed agreement proportion and $\bar{P}_e$ represents the expected chance agreement, defined respectively as:
\begin{equation}
\bar{P} = \frac{1}{N n (n-1)} \sum_{i=1}^N \sum_{j=1}^k n_{ij} (n_{ij} - 1), \quad \bar{P}_e = \sum_{j=1}^k p_j^2,
\end{equation}
where $p_j = \frac{1}{N n} \sum_{i=1}^N n_{ij}$.
Evaluated on this 1,000-pair subset, our annotation team achieves a score of $\kappa = 0.68 \pm 0.03$ (bounded by a 95\% confidence interval).
This systematic sampling guarantees that the Substantial Agreement reported is statistically robust.
Furthermore, we randomly inject 5\% honeypot pairs to test annotator attention, discarding data from any annotator who fails these checks.

\subsection{Human Preference Evaluation}
To validate the effectiveness of our proposed method, we conduct a Win/Tie/Lose evaluation against state-of-the-art baselines.
We employ five independent expert evaluators who are strictly distinct from the training annotation team to eliminate potential bias.
Evaluators view side-by-side, anonymized, and randomly ordered video pairs comparing our model against a baseline.
For each pair, they select one of three outcomes: Win (Ours is better), Tie (comparable quality), or Lose (Baseline is better).
The assessment requires evaluators to jointly weigh temporal consistency, prompt adherence, and overall perceptual quality.
To reduce subjective bias, a majority vote determines the final label for each comparison, requiring agreement from at least three of the five evaluators.
In cases where a majority consensus is not reached, a secondary review panel resolves the conflict.
This rigorous mechanism ensures that our reported win rates reflect a highly reliable human consensus.

\section{More Implementation Details}
\label{more-implementation-details}
\subsection{Training Setup}

Except for the open-source TurboDiffusion~\cite{zhang2025turbodiffusion}, we provide reproduction details for DMD2~\cite{yin2024improved}, DMDR~\cite{jiang2025distribution}, and FlashDMD~\cite{chen2025flash}.

\subsubsection{DMD2.}
The DMD2~\cite{yin2024improved} pipeline utilizes a few-step generator, a real model, and a fake model, all initialized from a pre-trained video diffusion backbone.
While the generator undergoes optimization, the real model remains frozen to provide a stable target distribution.
The framework minimizes the distribution matching loss by evaluating score gradient discrepancies between the real and fake models.
We follow the memory management configuration described in \cref{memory-management}.
We employ the AdamW optimizer with a learning rate of $1 \times 10^{-6}$ and weight decay of $0.01$.
The distillation targets a 4-step trajectory with a discrete timestep sequence $\tau \in \{1.000, 0.9544, 0.8748, 0.6997\}$ and a flow shift factor of $7$.
We train for $3,000$ steps with a global batch size of $4$ across 32 H20-96G GPUs.

\subsubsection{DMDR.} 

We adapt DMDR~\cite{jiang2025distribution} to the video domain by implementing its image-centric strategy atop the DMD~\cite{yin2023one} framework.
Beyond the standard distillation components, DMDR integrates the HPSv2 pixel-level reward model to enhance first-frame aesthetics and incorporates Dynamic LoRA modules into the frozen reference models.
During training, we keep only the LoRA modules and the generator trainable.
The objective linearly combines the DMD and ReFL losses.
For optimization, we use AdamW with $\beta=(0.9, 0.999)$, a weight decay of $1 \times 10^{-3}$, and a constant learning rate of $1 \times 10^{-6}$.
The distillation trajectory and hardware setup remain identical to those of DMD2.

\subsubsection{FlashDMD.} 
Following a similar cross-domain adaptation, FlashDMD~\cite{chen2025flash} is implemented atop the DMD2 architecture.
It augments the generator and reference models with a U-Net-based pixel discriminator for clean first-frame assessment and a latent reward model.
The training protocol partitions the flow trajectory, calculating the DMD loss in high-noise regions ($t \ge 0.9$) and adversarial loss in low-noise regions ($t < 0.9$).
After the primary distillation, we use the LRM to refine the generator's preference alignment.
We employ AdamW with $\beta=(0.5, 0.999)$ and a constant learning rate of $1 \times 10^{-6}$.
All other hyperparameters and computational costs remain consistent with the DMDR setup.

\input{tables_appdix/main_1}

\subsection{Memory Management}
\label{memory-management}
\input{tables_appdix/main_2}
\input{tables_appdix/main_3}
\input{tables_appdix/main_4}

\cref{tab_app_main_1} lists the memory strategy per model.

\section{More Experimental Results}
\label{more-experimental-results}
We show more experimental results as follows:
\begin{enumerate}
\item \textbf{Qualitative comparison with the Wan2.2 baseline}: We compare against the multi-step Wan2.2 base model under the 4-NFEs generation setting, as shown in \cref{fig_app:main_1}.
\item \textbf{Visualization results against competitive baselines}: We provide qualitative results comparing Reward Lightning with state-of-the-art methods across both 4-NFEs (\cref{fig_app:main_2}) and 1-NFEs (\cref{fig_app:main_3}) sampling regimes.
\item \textbf{Ablation analysis of BT and BTT losses}: In \cref{tab_app_main_2}, we evaluate the individual impacts of BT and BTT loss on the performance of our LRM.
\item \textbf{Investigation into training sequences}: In \cref{tab_app_main_3}, we study the optimization order, comparing sequential training (e.g., preference alignment followed by distillation, and vice versa) against our simultaneous homologous approach.
\item \textbf{Generalizability and reproducibility analysis}: In \cref{tab_app_main_4}, we extend our framework to HYVideo1.5~\cite{kong2024hunyuanvideo} to confirm its reproducibility and versatility across different generative backbones.
\end{enumerate}

\section{Limitation and Future Works}
\label{limitation-and-future-works}
\subsubsection{Limitations.}
Our framework is primarily constrained by the heterogeneous latent spaces across different video diffusion models.
Theoretically, a pixel-level reward model can decode any latent sample back into pixel space and evaluate it.
Thus, the pixel model is inherently generalizable and does not need to be retrained when the latent space of the VAE changes.
However, different VAEs compress markedly distinct latent spaces.
Consequently, a latent-level reward model specifically trained on a particular architecture (e.g., Wan2.1 or Wan2.2)~\cite{wan2025wan} cannot directly generalize to latent samples from generators utilizing different VAEs.
The latent incompatibility restricts the plug-and-play capability of our latent-based optimization across diverse generative families.

\subsubsection{Future Work.} 
To overcome this limitation, our future research focuses on two dimensions.
First, we aim to explore a train-free latent space conversion strategy to bridge the representation gap between different VAEs, thereby breaking the non-uniformity limitation and enabling cross-architecture reward evaluations.
Second, as multi-modal generation evolves, we plan to explore preference distillation for natively audio-video consistent models, extending human alignment from visual domains to synchronized multi-modal alignment.

\input{figures_appdix/main_1}
\input{figures_appdix/main_2}
\input{figures_appdix/main_3}

%% file: tables_appdix/main_1.tex
\begin{table}[t]
\centering
\caption{
\textbf{Memory management for experiments.}
\textbf{BF16} (bfloat16 Mixed-Precision), \textbf{GC} (Gradient Checkpointing),
\textbf{FSDP} (Fully Sharded Data Parallel with full shard), \textbf{SP} (Sequence Parallelism), 
\textbf{CO} (CPU Offload), \textbf{NG} (No Gradient)}.
\label{tab_app_main_1}
\begin{tabular}{ll}
\toprule
\textbf{Model Name} & \textbf{Memory Management Strategy} \\ \midrule
VAE \& Text Encoder & BF16, CO, SP=1, NG \\
Generator $v_\theta$ & BF16, GC, FSDP, SP=8 \\
EMA Generator $v_\theta^-$ & BF16, GC, FSDP, CO, NG \\
Reward Backbone $F_\phi$ & BF16, GC, FSDP, SP=8, NG \\
Reward Head $R_\phi^\text{head}$ & BF16, GC, FSDP, SP=1, NG \\
Discriminator Head $D_\psi^\text{head}$ & BF16, GC, FSDP, SP=1 \\
\bottomrule
\end{tabular}
\end{table}

%% file: tables_appdix/main_2.tex
\begin{table}[t]
\centering
\caption{
Ablation study on the reward objective.
}
\label{tab_app_main_2}
\resizebox{0.75\textwidth}{!}{%
\begin{tabular}{lcccc}
\toprule
\multirow{2}{*}{\textbf{Method}} & \multicolumn{2}{c}{\textbf{VideoGen-RewardBench}} & \multicolumn{2}{c}{\textbf{GenAI-Bench}} \\ \cmidrule(lr){2-3} \cmidrule(lr){4-5}
 & w/o Ties ($\uparrow$) & w/ Ties ($\uparrow$) & w/o Ties ($\uparrow$) & w/ Ties ($\uparrow$) \\ \midrule
LRM w/ BT  & 71.89 & 85.92 & 56.38 & 76.71 \\
LRM w/ BTT & 72.24 & 86.63 & 59.85 & 78.44 \\ \bottomrule
\end{tabular}
}
\end{table}

%% file: tables_appdix/main_3.tex
\begin{table}[t]
\centering
\caption{
Ablation study on the training sequence. 
}
\label{tab_app_main_3}
\resizebox{0.8\textwidth}{!}{%
\begin{tabular}{lcccc}
\toprule
\textbf{Training Sequence (4-NFEs)} & \textbf{TA ($\uparrow$)} & \textbf{MQ ($\uparrow$)} & \textbf{VQ ($\uparrow$)} & \textbf{I2V Score ($\uparrow$)} \\ \midrule
Distill $\rightarrow$ Preference Optimization & 92.15 & 69.45 & 69.12 & 92.59 \\
Preference Optimization $\rightarrow$ Distill & 89.34 & 71.87 & 69.34 & 93.17 \\
Ours & \textbf{96.32} & \textbf{75.39} & \textbf{69.78} & \textbf{95.34} \\ \bottomrule
\end{tabular}
}
\end{table}

%% file: tables_appdix/main_4.tex
\begin{table}[t]
\centering
\caption{
Generalizability and reproducibility analysis on HYVideo1.5-I2V.
}
\label{tab_app_main_4}
\resizebox{0.7\textwidth}{!}{%
\begin{tabular}{lcccc}
\toprule
\textbf{Methods} & \textbf{TA ($\uparrow$)} & \textbf{MQ ($\uparrow$)} & \textbf{VQ ($\uparrow$)} & \textbf{I2V Score ($\uparrow$)} \\ \midrule
\rowcolor{gray!20} \multicolumn{5}{l}{\textbf{80-NFEs}} \\
HYVideo1.5 & 95.51 & 74.23 & 67.18 & 92.48 \\
\rowcolor{gray!20} \multicolumn{5}{l}{\textbf{4-NFEs}} \\
HYVideo1.5 & 79.84 & 62.83 & 56.18 & 76.22 \\
DMDR (HYVideo1.5) & 95.29 & 71.78 & \textbf{69.24} & 92.98 \\
HPD (HYVideo1.5) & \textbf{96.11} & \textbf{75.10} & 67.89 & \textbf{95.17} \\ \bottomrule
\end{tabular}%
}
\end{table}

%% file: figures_appdix/main_1.tex
\begin{figure}[tb]
  \centering
  \includegraphics[width=\textwidth]{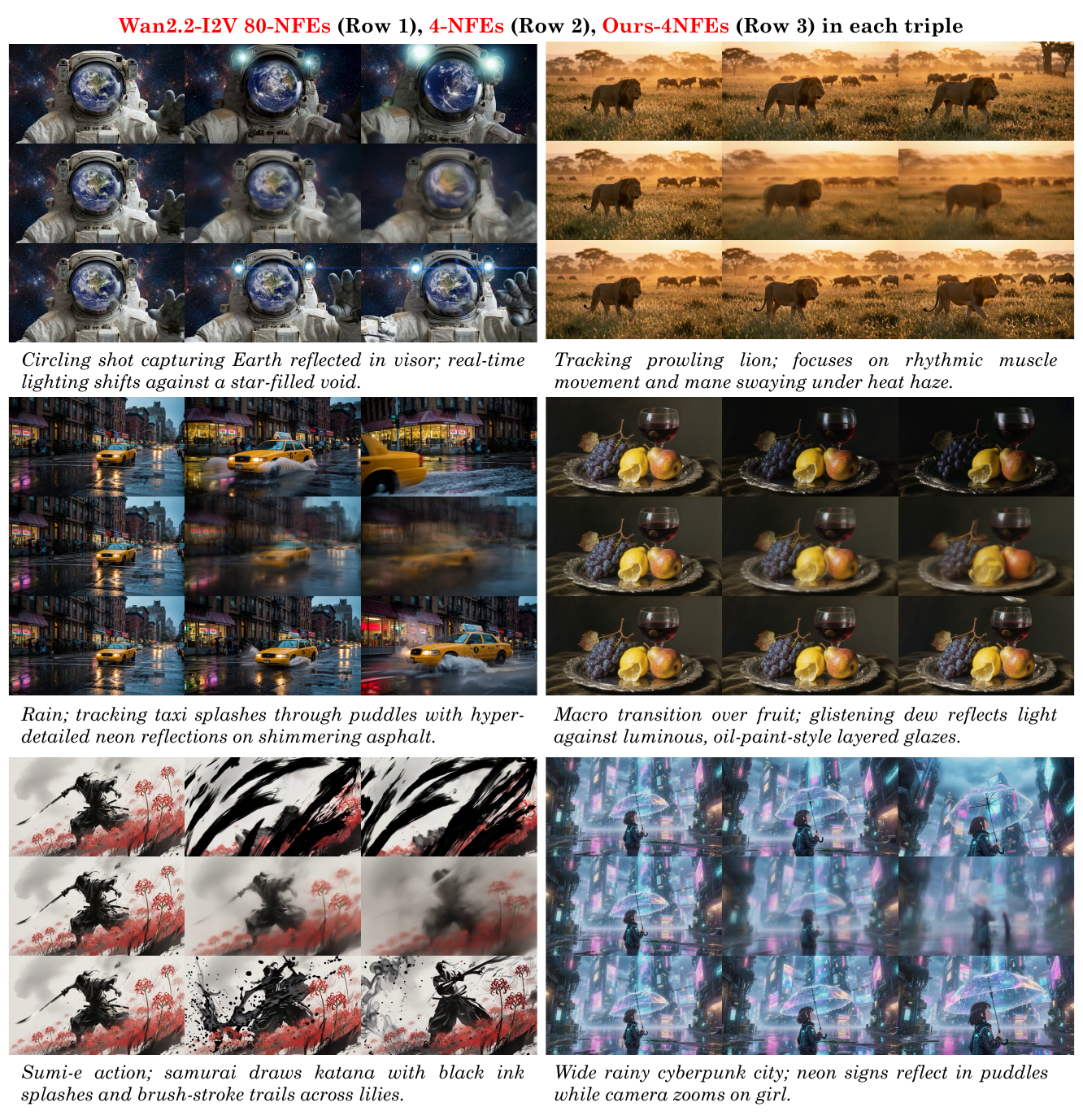}  
  \caption{
    \textbf{Qualitative Comparison.} In each triplet, rows 1 to 3 represent Wan2.2-I2V (80-NFEs), Wan2.2-I2V (4-NFEs), and our Reward Lightning (4-NFEs), respectively.
}
  \label{fig_app:main_1}
\end{figure}

%% file: figures_appdix/main_2.tex
\begin{figure}[tb]
  \centering
  \includegraphics[width=\textwidth]{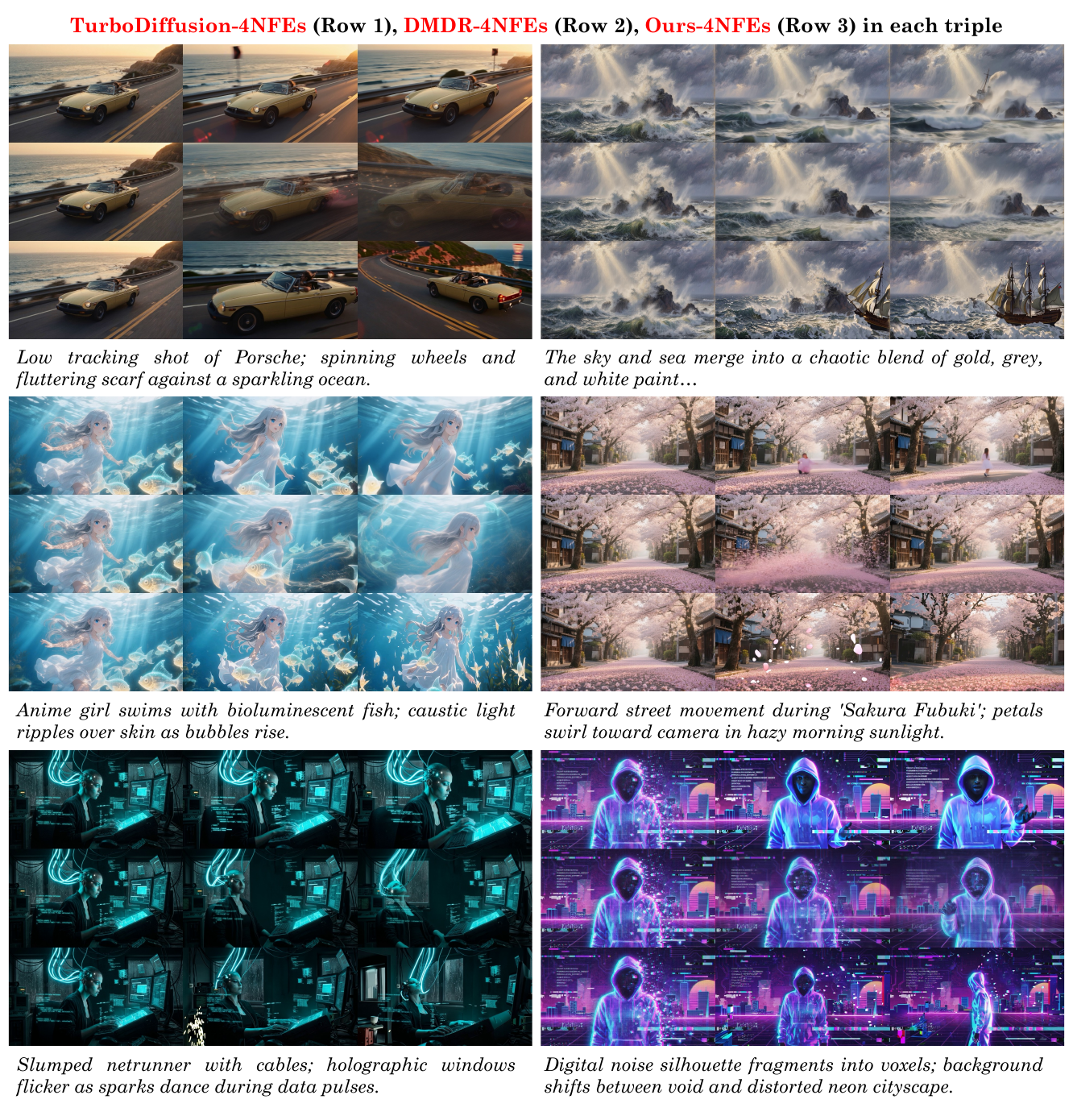}  
  \caption{
    \textbf{Qualitative Comparison.} In each triplet, rows 1 to 3 represent TurboDiffusion (4-NFEs), DMDR (4-NFEs), and our Reward Lightning (4-NFEs), respectively.
}
  \label{fig_app:main_2}
\end{figure}

%% file: figures_appdix/main_3.tex
\begin{figure}[tb]
  \centering
  \includegraphics[width=\textwidth]{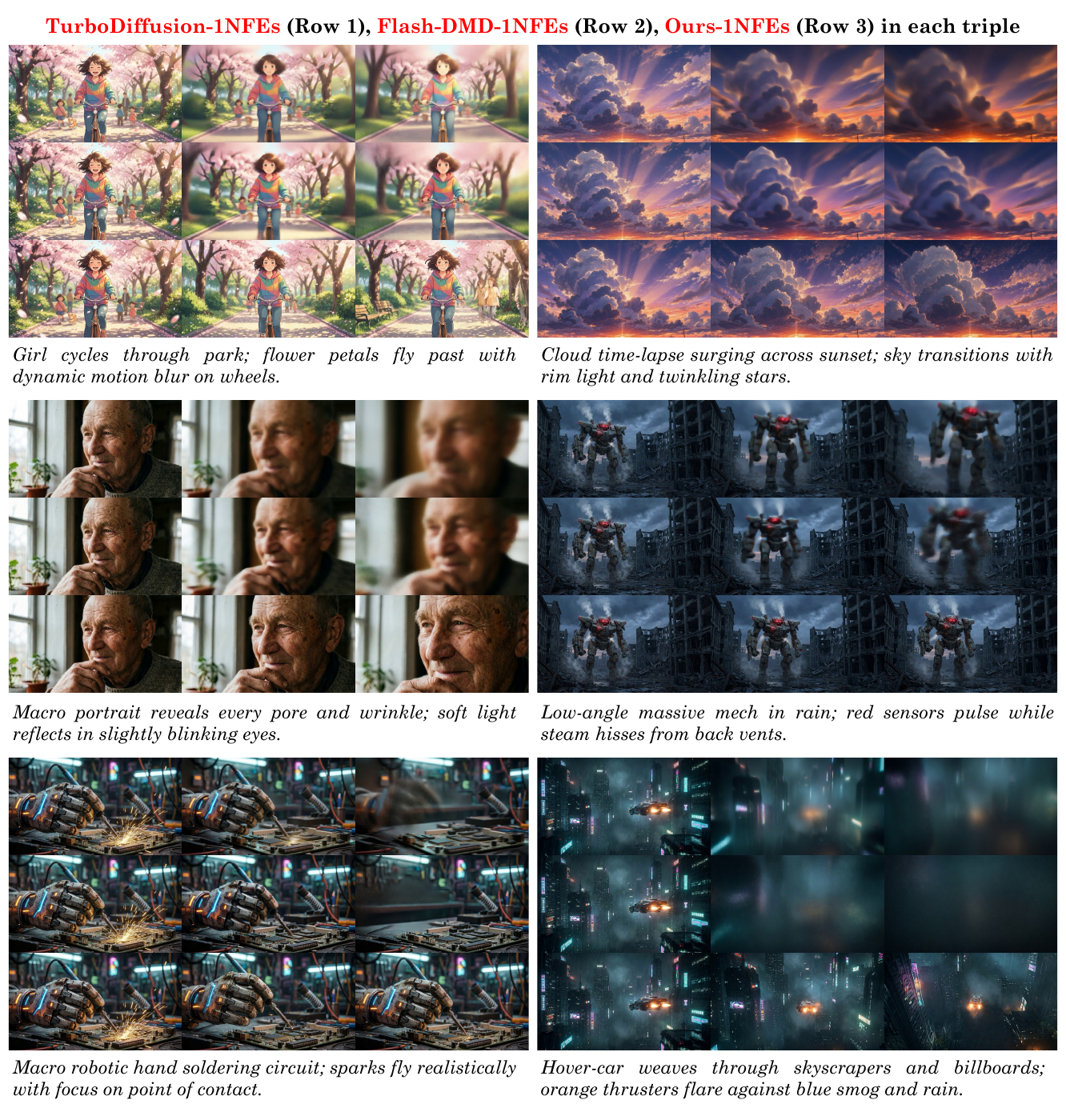}  
  \caption{
    \textbf{Qualitative Comparison.} In each triplet, rows 1 to 3 represent TurboDiffusion (1-NFEs), FlashDMD (1-NFEs), and our Reward Lightning (1-NFEs), respectively.
}
  \label{fig_app:main_3}
\end{figure}